\RequirePackage{fix-cm}
\documentclass[twocolumn]{svjour3}           %
\smartqed  %
\usepackage{graphicx}
\usepackage{url}              %
\usepackage{booktabs}         %
\usepackage{amsfonts} 
\usepackage{amsmath}
\usepackage{amssymb}%
\usepackage{color}            %
\usepackage{xcolor}    %
\usepackage{multirow}
\usepackage{adjustbox}
\usepackage{caption} %
\usepackage[misc]{ifsym}
\usepackage[pagebackref=true,breaklinks=true,letterpaper=true,colorlinks,bookmarks=false]{hyperref}
\usepackage[labelfont=bf]{caption}
\usepackage{colortbl}
\usepackage{tabularx}
\usepackage{tikz,dcolumn,subcaption}
\usepackage{comment}
\usepackage{algorithmicx}
\usepackage{algorithm}
\usepackage{algpseudocode} 
\definecolor{turquoise}{cmyk}{0.65,0,0.1,0.3}
\definecolor{purple}{rgb}{0.65,0,0.65}
\definecolor{dark_green}{rgb}{0, 0.5, 0}
\definecolor{orange}{rgb}{0.8, 0.6, 0.2}
\definecolor{red}{rgb}{0.8, 0.2, 0.2}
\definecolor{darkred}{rgb}{0.6, 0.1, 0.05}
\definecolor{blueish}{rgb}{0.0, 0.3, .6}
\definecolor{light_gray}{rgb}{0.8, 0.8, 0.8}
\definecolor{pink}{rgb}{1, 0, 1}
\definecolor{greyblue}{rgb}{0.25, 0.25, 1}
\definecolor{mistyrose}{rgb}{1.0, 0.89, 0.88}
\definecolor{whitee}{rgb}{1.0, 1.0, 1.0}
\definecolor{palerobineggblue}{rgb}{0.59, 0.87, 0.82}
\definecolor{lavenderblue}{rgb}{0.9, 0.9, 1.0}

\definecolor{darkblue}{HTML}{6082B6}
\definecolor{darkorange}{HTML}{FF8C00}

\newcommand{\roland}[1]{{\color{red}#1}}

\usepackage[english]{babel}
\begin{document}

\title{PUFM++: Point Cloud Upsampling via Enhanced Flow Matching}

\author{Zhi-Song Liu$^{1}$, Chenhang He$^{2}$, Roland Maier$^{3}$, Andreas Rupp$^{4}$}
\authorrunning{Liu et al.}

\institute{
    $^1$ Department of Computational Engineering, Lappeenranta-Lahti University of Technology LUT, Finland \\
    $^2$ The Hong Kong Polytechnic University \\
    $^3$ Institute for Applied and Numerical Mathematics at Karlsruhe Institute of Technology (KIT), Germany \\
    $^4$ Department of Mathematics at Saarland University, Germany
}

\date{Received: XXX / Accepted: XXX}

\maketitle

\abstract{
Recent advances in generative modeling have demonstrated strong promise for high-quality point cloud upsampling. In this work, we present PUFM++, an enhanced flow-matching framework for reconstructing dense and accurate point clouds from sparse, noisy, and partial observations. PUFM++ improves flow matching along three key axes: (i)~geometric fidelity, (ii)~robustness to imperfect input, and (iii)~consistency with downstream surface-based tasks.
We introduce a two-stage flow-matching strategy that first learns a direct, straight-path flow from sparse inputs to dense targets, and then refines it using noise-perturbed samples to approximate the terminal marginal distribution better. To accelerate and stabilize inference, we propose a data-driven adaptive time scheduler that improves sampling efficiency based on interpolation behavior. We further impose on-manifold constraints during sampling to ensure that generated points remain aligned with the underlying surface. Finally, we incorporate a recurrent interface network~(RIN) to strengthen hierarchical feature interactions and boost reconstruction quality.
Extensive experiments on synthetic benchmarks and real-world scans show that PUFM++ sets a new state of the art in point cloud upsampling, delivering superior visual fidelity and quantitative accuracy across a wide range of tasks. Code and pretrained models are publicly available at {\small{\url{https://github.com/Holmes-Alan/Enhanced_PUFM}}}. 


}

\maketitle

\section{Introduction}
In 3D data processing, point clouds are a fundamental yet versatile data structure, widely adopted in robotic navigation, industrial inspection, autonomous driving, and 3D scene reconstruction. Their unstructured nature enables efficient acquisition and processing; however, many downstream tasks, such as object recognition~\cite{robot}, surface reconstruction~\cite{p2m}, and semantic scene understanding~\cite{car}, require dense, uniformly distributed, and noise-free point clouds. Generating such high-quality data typically requires expensive, high-resolution 3D acquisition devices, including LiDAR systems and depth cameras.

To overcome the limitations of hardware-centric approaches, recent advances in deep learning for 3D geometry have spurred interest in learning-based point cloud upsampling~\cite{pugeo,pu-dense,puflow}, aiming to infer dense outputs from sparse inputs statistically. This task is conceptually related to image super-resolution~\cite{sr_1,sr_2}, yet introduces unique challenges due to the irregular, unordered nature of point sets, the presence of noise, and the prevalence of incomplete or topologically ambiguous shapes.

Traditional methods often formulate point cloud upsampling as an optimization problem, relying heavily on handcrafted geometric priors and deterministic rules. While effective to some extent, these approaches are generally limited in terms of generalization and scalability. The introduction of PUNet~\cite{punet} marked a significant shift towards end-to-end learning, enabling the network to infer high-resolution point distributions directly from coarse inputs. Subsequent research further refined this line of work by modeling local geometric patches, thereby exploiting the manifold structure of point clouds and learning recurring patterns across spatial scales~\cite{mpu}. Despite these advances, the lack of expressive feature representations and globally consistent optimization has motivated the development of generative learning frameworks~\cite{pugan,puflow,pudm,pufm}, which seek to model complex point distributions more holistically.

Among these, diffusion probabilistic models~\cite{ddpm} and flow-matching-based methods~\cite{flow} have emerged as promising generative paradigms for 3D data synthesis. PUDM~\cite{pudm} and PUFM~\cite{pufm} are recent state-of-the-art techniques that leverage these principles for point cloud upsampling, showing strong performance across benchmarks. While diffusion models simulate a stochastic denoising process to generate data samples, flow matching provides a more direct and computationally efficient alternative by learning a continuous mapping from sparse to dense point sets.

In this work, we present PUFM++, a scalable flow-matching architecture designed for arbitrary point cloud upsampling across diverse object categories and geometric scales. Building upon our earlier framework PUFM~\cite{pufm}, PUFM++ performs flow matching in the local patch space, enabling a more flexible and fine-grained modeling of geometric structures. Notably, PUFM++ achieves high-quality upsampling with as few as five inference steps, offering substantial improvements in efficiency while maintaining strong fidelity. Extensive experiments on multiple datasets demonstrate that PUFM++ achieves state-of-the-art performance. The analysis on different resolutions and real-world scenarios validates the superior performance, efficiency, and generalization to real-world examples. 

Our contributions are summarized as follows: 
\begin{itemize}
    \item \textbf{Two-Stage Flow Matching Strategy.} 
    We introduce a two-stage training framework for flow matching. The first stage optimizes the full transport trajectory between sparse and dense point distributions, while the second stage applies a dedicated refinement step that enhances initial inference quality, improving both fidelity and robustness.

    \item \textbf{Adaptive Time Sampling.} 
    To accelerate inference, we propose an adaptive sampling scheme based on discretized ordinary differential equations (ODEs) that adjusts integration step sizes according to statistics derived from training trajectories, substantially reducing the number of sampling steps.

    \item \textbf{Test-Time Optimization and Post-Processing.} 
    We incorporate a manifold-aware prior that encourages smooth transport trajectories, and apply a lightweight $k$NN-based post-processing module to improve local geometric consistency, leading to higher-quality downstream mesh reconstruction.

    \item \textbf{Latent-State Recurrent Velocity Estimation.} 
    To preserve long-range geometric consistency throughout the continuous reverse process, we extend the standard velocity network to a recurrent formulation with a compact evolving latent state. This memory mechanism provides persistent global shape context across noise levels, yielding marked improvements in fidelity, density uniformity, and structure preservation.
\end{itemize}


    
    

The basis of this work, PUFM, appeared at AAAI 2026~\cite{pufm}. PUFM++ represents a major extension of PUFM in three primary aspects. (1) On the optimization side, PUFM++ integrates a two-stage flow matching scheme, statistics-guided time step sampling, and a manifold-aware prior, collectively addressing the convergence limitations of PUFM and substantially improving both efficiency and reconstruction fidelity (Sections 3.2.1–3.2.3). (2) Architecturally, PUFM++ replaces the original PointNet++ backbone with a recurrent latent self-conditioning mechanism inspired by Recurrent Interface Networks (RINs), enabling iterative refinement and improved spatial coherence, which are capabilities absent in the feedforward design of PUFM (Section 3.3). (3) Finally, we provide a significantly expanded empirical study, including more datasets, higher resolutions, and extensive real-world evaluations, demonstrating consistent improvements in fidelity, robustness, and generalization (Section 4).

\section{Related Work}
\label{sec:relwork}
\subsection{Point clouds and their analysis}
Different from image data, point clouds contain irregular structures and are invariant to permutations. PointNet~\cite{pointnet} is the pioneering work that applies shared MLPs to extract point-wise features. PointNet++~\cite{pointnet++} significantly improves upon it by proposing a hierarchical architecture with set abstraction layers to extract multi-level features. This encoder-decoder framework has become a backbone for many subsequent tasks, including upsampling. Subsequent research has expanded in several directions. Graph-based methods like DGCNN~\cite{dgcnn} dynamically construct a graph to capture local structures. Convolution-based methods, such as PointConv~\cite{pointconv} and KPConv~\cite{kpconv}, define continuous or discrete convolutions directly on point clouds. Voxel-based hybrids like PVCNN~\cite{pvcn} combine the efficiency of sparse voxel convolutions with the precision of point-based networks to improve computational efficiency. More recently, transformer-based architectures have demonstrated impressive performance. Models like Point Transformer~\cite{point} and its more efficient successor Point Transformer V2~\cite{point_v2} use vector self-attention to model long-range dependencies. Point Transformer V3~\cite{point_v3} further improves the efficiency for multiple downstream tasks by studying the trade-offs between accuracy and efficiency. Most recently, the DeepLA-Net~\cite{deepla} framework further popularized the use of deep neural networks for point cloud processing, surpassing SOTA performance on multiple tasks.

\subsection{Learning-based point cloud upsampling}
Point cloud upsampling aims to generate dense, uniform point sets that faithfully represent the underlying surface. Optimization-based and early learning-based methods often relied on handcrafted priors. The advent of deep learning led to pioneering data-driven works such as PU-Net~\cite{punet}, which uses a feature-expansion module followed by multi-layer perceptrons. EC-Net~\cite{ecnet} introduced edge-aware training for better feature preservation. MPU~\cite{mpu} proposed a multi-patch progressive network to handle complex surfaces. To better model local geometry, PU-GCN~\cite{pu-gcn} introduced a graph convolutional network with a NodeShuffle upsampling module. PUGAN~\cite{pugan} was the first to employ a generative adversarial network (GAN)~\cite{gan} to encourage the generation of a uniform point distribution, mitigating the issue of clustered points. Dis-PU~\cite{dispu} later improved upon this by disentangling the upsampling and refinement stages. Another line of work explicitly models the underlying surface. PUGeo-Net~\cite{pugeo} learns a geodesic mapping to project points onto a 2D parameterized domain for upsampling. Similarly, NePs~\cite{np} treats upsampling as a neural implicit surface reconstruction problem. SPU-Net~\cite{spunet} leverages self-attention and a learnable seed generation module to capture both global and local structures. PU-Transformer~\cite{putransformer} expanded attention mechanisms to the transformer-based designs to capture long-range geometric dependencies. Recognizing that real-world point clouds are often noisy, recent methods have begun to integrate denoising. SUperPC~\cite{superpc} and PU-Dense~\cite{pu-dense} jointly perform upsampling and denoising, showing that the two tasks can be mutually beneficial. Despite these advances, a significant limitation of most prior work is the heavy reliance on the Chamfer Distance (CD) as the primary supervision signal. While computationally efficient, CD is known to be insensitive to the underlying distribution and can lead to biased or non-uniform point distributions, failing to capture fine geometric details.

\subsection{Diffusion model and flow matching}
Generative models, particularly diffusion models~\cite{ddpm} and flow matching methods~\cite{flow}, have recently achieved remarkable success in point cloud generation and processing. Diffusion model-based methods learn to denoise a Gaussian distribution into a data distribution. Diffusion Point Cloud~\cite{pdm} and PVD~\cite{pvd} are pioneering works that applied diffusion models to 3D point cloud generation and completion. GFNet~\cite{GFNet} introduced flow matching for 3D shape synthesis, aligning source and target distributions through optimal transport trajectories learned from data. For upsampling, PUDM~\cite{pudm} frames the task as a conditional generation problem, where a sparse input guides the reverse diffusion process to generate a dense output. PDANS~\cite{pdans} further improves it by introducing the adaptive noise suppression (ANS) module to upsample point clouds against noise robustly. On the one hand, these methods often require $20$ to $50$ denoising steps, which slows inference. On the other hand, Normalizing Flows~\cite{normflow} learn an invertible mapping between a simple prior and a complex data distribution. PUFlow~\cite{puflow} is a seminal work that represents a 3D shape as an invertible flow over a latent space for point cloud upsampling. More recently, Flow Matching~\cite{flow} has emerged as a simpler, more efficient alternative to diffusion models, offering a simulation-free training objective that achieves superior performance with fewer sampling steps. While applied to point cloud generation (PCFlow~\cite{pcflow}), its potential for conditional tasks, such as upsampling, remains largely unexplored. One recent work is PUFM~\cite{pufm}, which learn a straight flow between sparse and dense point clouds for upsampling. Due to its deterministic trajectory-based learning, it can efficiently upsample point clouds with as few as five steps.

\section{Method}
\label{sec:method}
\subsection{Preliminary}

Flow matching (FM)~\cite{flow} and its extension, conditional flow matching (CFM)~\cite{cfm}, are effective frameworks for learning transport dynamics between probability distributions via ordinary differential equation (ODE) flows. The goal of FM is to learn a time-dependent vector field \( \nu_\theta(\boldsymbol{x}, t) \), parameterized by a neural network with parameters $\theta$, whose induced flow transports a source distribution \( \rho_0 \) to a target distribution \( \rho_1 \). The resulting probability density path \( p_t(\boldsymbol{x}) \) satisfies the continuity equation
\begin{equation}
\frac{\partial p_t(\boldsymbol{x})}{\partial t}
+ \nabla \cdot \left( p_t(\boldsymbol{x}) \nu_\theta(\boldsymbol{x}, t) \right) = 0.
\label{eq:fm}
\end{equation}
Directly learning the marginal vector field \( \nu_\theta(\boldsymbol{x}, t) \) is generally intractable. CFM addresses this challenge by introducing a conditional probability path \( p_t(\boldsymbol{x} \mid \boldsymbol{z}) \) and a corresponding conditional vector field \( u_t(\boldsymbol{x} \mid \boldsymbol{z}) \), where \( \boldsymbol{z} \) denotes conditioning variables, typically the data endpoints. A commonly used conditional path is a Gaussian distribution whose mean follows a linear interpolation between the endpoints. That is,
\begin{align}
p_t(\boldsymbol{x} \mid \boldsymbol{x}_0, \boldsymbol{x}_1)
&= \mathcal{N}\!\left(
\boldsymbol{x} \,\middle|\, \mu_t(\boldsymbol{x}_0, \boldsymbol{x}_1),
\, \sigma_t^2 \boldsymbol{I}
\right), \label{eq:cond_prob} \\
\mu_t(\boldsymbol{x}_0, \boldsymbol{x}_1)
&= t \boldsymbol{x}_1 + (1 - t) \boldsymbol{x}_0,
\label{eq:line}
\end{align}
where \( \boldsymbol{x}_0 \sim \rho_0 \) and \( \boldsymbol{x}_1 \sim \rho_1 \). The conditional vector field \( u_t(\boldsymbol{x} \mid \boldsymbol{x}_0, \boldsymbol{x}_1) \) is defined such that its induced flow recovers this prescribed conditional path. The network parameters \( \theta \) are learned by minimizing the CFM objective, which regresses the learned vector field \( \nu_\theta \) onto the conditional vector field. That is, we minimize
\begin{equation}
\mathbb{E}_{t, \boldsymbol{x}_0, \boldsymbol{x}_1, \boldsymbol{x}_t}
\left[
\left\|
\nu_\theta(\boldsymbol{x}_t, t)
-
u_t(\boldsymbol{x}_t \mid \boldsymbol{x}_0, \boldsymbol{x}_1)
\right\|_2^2
\right],
\label{eq:fm_2}
\end{equation}
where \( t \sim \mathcal{U}[0,1] \), the endpoint pairs
\( (\boldsymbol{x}_0, \boldsymbol{x}_1) \sim \pi(\boldsymbol{x}_0, \boldsymbol{x}_1) \)
are sampled from a coupling \( \pi \) between \( \rho_0 \) and \( \rho_1 \), and
\( \boldsymbol{x}_t \sim p_t(\cdot \mid \boldsymbol{x}_0, \boldsymbol{x}_1) \). In the special case of optimal transport conditional flow matching (OT-CFM), the conditional path is noise-free, corresponding to the limit \( \sigma_t \to 0 \). This results in a deterministic straight-line trajectory
\(
\boldsymbol{x}_t = t \boldsymbol{x}_1 + (1 - t) \boldsymbol{x}_0
\),
and the associated conditional vector field becomes time-independent for a fixed endpoint pair, i.e.,
\(
u_t(\boldsymbol{x}_t \mid \boldsymbol{x}_0, \boldsymbol{x}_1)
= \boldsymbol{x}_1 - \boldsymbol{x}_0
\).

During inference, the learned flow is approximated using a discrete-time Euler integrator. Given a step size~\( \tau \), the state is updated as
\begin{equation}
\boldsymbol{x}_{t+\tau}
=
\boldsymbol{x}_t
+
\tau \, \nu_\theta(\boldsymbol{x}_t, t).
\label{eq:fm_3}
\end{equation}

Despite their success, FM and CFM share limitations similar to those discussed in~\cite{sb_1,sb_2}. First, standard FM formulations yield deterministic ODE trajectories and therefore cannot recover the stochastic dynamics required for Schr\"odinger Bridge problems. Second, while linear CFM provides an approximation to optimal transport displacement, it does not strictly solve the entropic optimal transport problem between \( \rho_0 \) and \( \rho_1 \) without carefully designed couplings \( \pi \).
\begin{figure}[t]
	\centering
		\centerline{\includegraphics[width=0.5\textwidth]{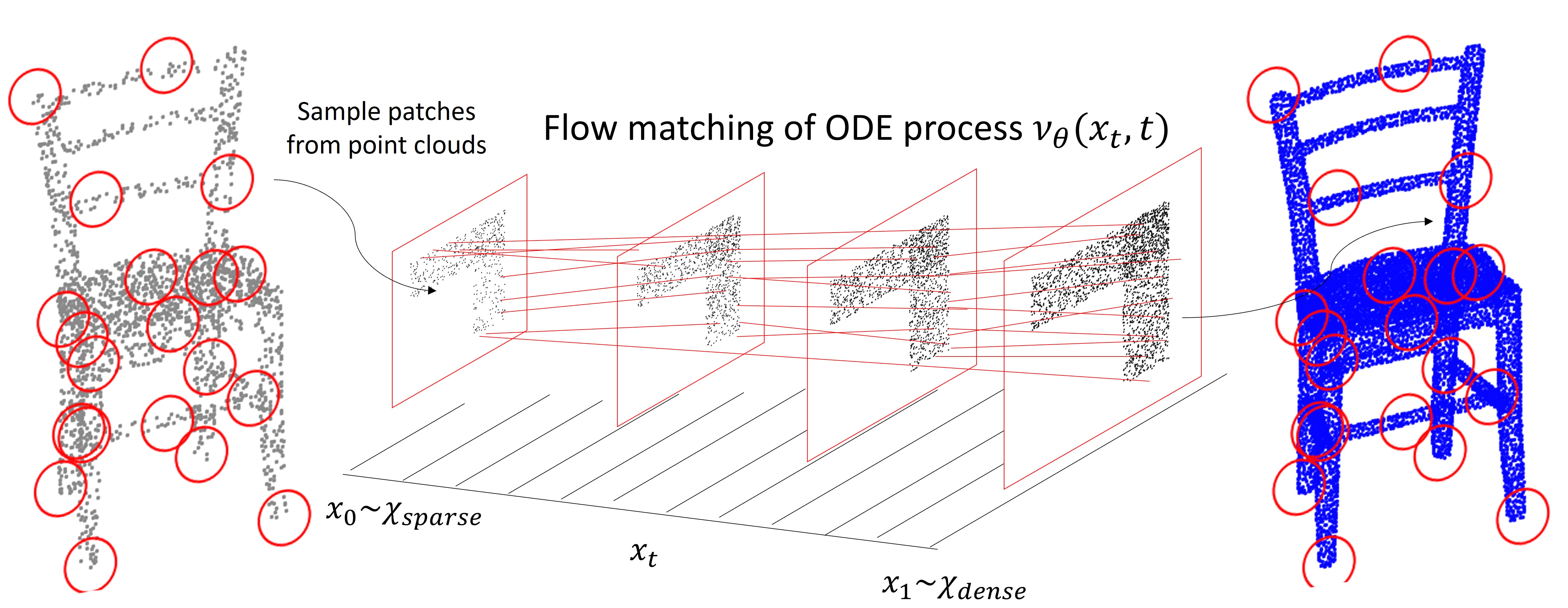}}
		\caption{\textbf{Flow Matching for Point Cloud Upsampling.} We extract paired patches from sparse and dense point clouds and learn the velocity field for point cloud upsampling.
		}
		\label{fig:overall_fm}
\end{figure}

\subsection{Patch-based Point Cloud Upsampling via Flow Matching}
We propose a patch-based point cloud upsampling via enhanced Flow Matching (PUFM++), mainly inspired by PUDM~\cite{pudm} and PUFM~\cite{pufm}. The former one learns a conditional diffusion model for point cloud upsampling, i.e., a mapping from noise to data. Thus, it is slow and thereby suboptimal. Contrarily, PUFM proposes to use flow matching to learn a sparse-to-dense distribution mapping and achieve an acceleration of more than factor 10. In this paper, we extend the work on PUFM and propose new techniques to improve it further, as explained in the following.

\subsubsection{Two-stage flow matching optimization}
Assume that we are given a dense point cloud $\mathcal X_{\text{dense}}\in \mathbb{R}^{N\times3}$ with $N$ points and its sparse version $\mathcal X_{\text{sparse}}\in \mathbb{R}^{M\times3}$ with $M$ points, where $M \ll N$. We initially densify the sparse point cloud via a midpoint interpolation~\cite{grad-pu} method and obtain the naively upsampled point cloud $\mathcal X'_{\text{dense}}\in \mathbb{R}^{M\times3}$. Then we extract corresponding patch pairs as $\boldsymbol{x}_0\sim \mathcal X'_{\text{dense}}\in \mathbb{R}^{Q\times3}$ and $\boldsymbol{x}_1\sim \mathcal X_{\text{dense}}\in \mathbb{R}^{Q\times3}$. As shown in Figure~\ref{fig:overall_fm}, our goal is to learn a flow matching model in the patch space by finding the optimal transport between $\boldsymbol{x}_0$ and $\boldsymbol{x}_1$. 

\noindent \textbf{Stage 1: Pre-aligned Flow Matching Optimization.} Based on \eqref{eq:fm_2}, we learn a flow matching model to estimate the velocity field as $\nu_\theta(\boldsymbol{x}_t, t)$. In most cases, we assume that the underlying OT plan $\pi(\boldsymbol{x}_0, \boldsymbol{x}_1)$ is a pointwise correspondence between sparse and dense point clouds. Hence, we define the interpolants as in~\eqref{eq:line}, 
where $\boldsymbol{x}_1$ and $\boldsymbol{x}_0$ are sampled from the training patches \( \mathcal X'_{\text{dense}} \) and \( \mathcal X_{\text{dense}} \), respectively. However, point clouds are unordered in 3D space and are irregularly distributed. The true paired point patches should be sampled from the OT plan as pair $(\boldsymbol{x}_0, \boldsymbol{x}_1)\sim \pi^*$, which minimizes the expected cost of transporting mass from the source to the target domain. That is,
\begin{equation}
\pi^* = \arg\min_{\pi\in \Pi(\boldsymbol{x}_0, \boldsymbol{x}_1)} \int ||\boldsymbol{x}_1-\boldsymbol{x}_0||^2 \,\mathrm{d}\pi (\boldsymbol{x}_0, \boldsymbol{x}_1)
\label{eq:fm_6}
\end{equation}
where \( \Pi(\boldsymbol{x}_0, \boldsymbol{x}_1) \) is the set of all couplings (joint distributions) with margins of $\boldsymbol{x}_0$ and $\boldsymbol{x}_1$. 

To find the OT plan, we utilize the Wasserstein distance to measure the source and target distributions as $\rho_0=\sum_{i=1}^N a_i\delta_{\boldsymbol{x}_0}$ and $\rho_1=\sum_{i=1}^N b_i\delta_{\boldsymbol{x}_1}$, where $\delta_{\boldsymbol{x}_i}$ refers to the Dirac delta distribution centered at the point $\boldsymbol{x}_i$. The earth mover's distance (EMD), i.e., the Wasserstein 1-distance, is given by
\begin{equation}
\begin{aligned}
\operatorname{EMD}(\rho_0,&\rho_1)\\&=\min_{\pi\in\Pi(\boldsymbol{x}_0, \boldsymbol{x}_1)} \sum_{i=1}^{N}\sum_{j=1}^M \pi_{ij}||\boldsymbol{x}_0(i)-\boldsymbol{x}_1(j)||
\end{aligned}
\label{eq:emd}
\end{equation}
where $x_0(i), x_1(j)$ represent the sparse and dense sampled points, $\sum_j \pi_{i,j}=a_i$ and $\sum_i \pi_{i,j}=b_j$. We then get a point-to-point correspondence between the sparse and dense point clouds. Notably, a major challenge of using EMD is its computational cost; PointMixup~\cite{emd} proposes an EMD approximation via an auction algorithm. It aligns two point clouds by treating source points as bidders and target points as items, iteratively raising “prices” and reassigning matches so that each source gets its most cost-effective target, yielding an $\epsilon$-approximate optimal transport plan $\pi^\epsilon$ in $\mathcal{O}(n^2)$ parallel time. That is, $\pi^\epsilon\in \Pi(\rho_0, \rho_1)$ with
\begin{equation}
\sum_{i,j}c_{i,j}\pi^\epsilon_{i,j} \leq \sum_{i,j}c_{i,j}\pi_{i,j} + \epsilon.
\label{eq:emd_appro}
\end{equation}
Here, $c_{i,j}=||\boldsymbol{x}_1(i) - \boldsymbol{x}_0(j)||^2$ is the cost of matching $\boldsymbol{x}^i_1$ to $\boldsymbol{x}^j_0$. Utilizing this approximated EMD, we can find the matching function $\varphi\colon \{1,\dots,N\}\to\{1,\dots,M\}$ that assigns each dense point $\boldsymbol{x}_1$ to the sparse point $\boldsymbol{x}_0$. Then we train a pre-aligned flow matching based on minimizing
\begin{equation}
\mathcal{L}_\text{MSE} = \mathbb{E}_{t\sim\mathcal{U}[0,1]}||\nu_\theta(\boldsymbol{x}_t, t)-(\boldsymbol{x}_1 - \boldsymbol{x}_0)||^2. 
\label{eq:loss_1}
\end{equation}
where $t$ is sampled from the uniform distribution $\mathcal{U}[0,1]$.

\textbf{Stage 2: Endpoint Flow Matching Refinement.} We propose a second-step refinement on the pre-aligned flow matching model to boost the quality of point cloud upsampling further. In~\eqref{eq:emd_appro}, we perform OT-guided flow matching by regressing velocities along displacement paths built from an (approximate) optimal assignment $\pi^\epsilon$, which teaches locally consistent transport directions. It does not guarantee that the global pushforward matches $\rho_1$ once we integrate from only $\boldsymbol{x}_0$ (no access to $\boldsymbol{x}_1$ at the sampling time), nor that the density is correct where OT pairings were noisy or sparse. 

In the second step, we integrate trajectories from the source only and minimize a permutation-invariant chamfer distance (CD) loss between the sparse and dense point clouds. Furthermore, we add additional noise to the sparse point clouds to make the learned flow robust to input perturbations and encourage local coherence of trajectories. The desired value is then given by 
\begin{multline}
\mathcal{L}_\text{CD} = \frac{1}{\|\Tilde{\boldsymbol{x}}_1\|}\sum_{\Tilde{\boldsymbol{x}}_1(i)\in \Tilde{\boldsymbol{x}}_1} \min_{\boldsymbol{x}_1(i)\in \boldsymbol{x}_1} ||\Tilde{\boldsymbol{x}}_1(i)-\boldsymbol{x}_1(i)||^2 \\
+ \frac{1}{\|\boldsymbol{x}_1\|}\sum_{\boldsymbol{x}_1(i)\in \boldsymbol{x}_1} \min_{\Tilde{\boldsymbol{x}}_1(i)\in \Tilde{\boldsymbol{x}}_1} ||\boldsymbol{x}_1(i)-\Tilde{\boldsymbol{x}}_1(i)||^2,
\label{eq:loss_2}
\end{multline}
where $\Tilde{\boldsymbol{x}}_1=\nu_\theta(\boldsymbol{x}_0+\xi,\ 0)$ with $\xi\sim\mathcal{N}(0,\sigma^2I)$. This enforces the terminal marginal constraints, reminiscent of the Schr\"odinger system, adjusting the potentials so that the learned distribution $\rho^\theta_1$ approximately equals the true $\rho_1$. More importantly, the additional noise $\xi$ introduces a smoothing regularizer into the flow-matching model. It reduces the gradient variance across nearby initial sparse points and pushes the learned field to transform the neighborhoods of $\boldsymbol{x}_0(i)$ toward $\boldsymbol{x}_1(i)$ consistently. The summary is shown in Algorithm \textcolor{red}{1}.

\begin{algorithm}[t]
\caption{Two-Stage Flow Matching for Point Cloud Upsampling}
\begin{algorithmic}
\Require $\mathcal{X}_{\text{sparse}}$, $\mathcal{X}_{\text{dense}}$, noise scale $\sigma$
\State \textbf{Stage 1: Pre-aligned Flow Matching}
\For{each training iteration}
    \State $\boldsymbol{x}_0 \sim \mathcal{X}_{\text{sparse}}, \quad \boldsymbol{x}_1 \sim \mathcal{X}_{\text{dense}}$
    \State $t \gets 1 - \cos(s\pi/2), \quad s \sim \mathcal{U}[0,1]$
    \State $\Tilde{\boldsymbol{x}}_0 \gets \text{mid}(\boldsymbol{x}_0, \eta)$ \Comment{midpoint interpolation}
    \State $\varphi^* \gets \pi(\Tilde{\mathbf{x}}_0, \boldsymbol{x}_1)$ \Comment{OT pre-alignment (auction)}
    \State $\boldsymbol{x}_t \gets (1-t)\Tilde{\boldsymbol{x}}_0 + t \boldsymbol{x}_1^{\varphi^*(i)}$
    \State $\mathcal{L}_{\text{MSE}} \gets \|\nu_\theta(\boldsymbol{x}_t, t) - (\boldsymbol{x}_1^{\varphi^*(i)} - \Tilde{\boldsymbol{x}}_0)\|_2^2$
    \State Update $\theta$ using $\mathcal{L}_{\text{FM}}$
\EndFor
\vspace{0.5em}
\State \textbf{Stage 2: Endpoint Flow Matching Refinement}
\For{each refinement iteration}
    \State $\boldsymbol{x}_0 \sim \mathcal{X}_{\text{sparse}}, \quad \boldsymbol{x}_1 \sim \mathcal{X}_{\text{dense}}$
    \State $\boldsymbol{x}_0' \gets \boldsymbol{x}_0 + \sigma \,\xi, \quad \xi \sim \mathcal{N}(0, I)$ \Comment{additional noise}
    \State $\boldsymbol{x}_{\text{pred}} \gets \boldsymbol{x}_0' + \nu_\theta(\boldsymbol{x}_0', 0)$
    \State $\mathcal{L}_\mathrm{CD} \gets \text{CD}(\boldsymbol{x}_{\text{pred}}, \boldsymbol{x}_1)$
    \State Update $\theta$ using $\mathcal{L}_\mathrm{CD}$
\EndFor
\end{algorithmic}
\end{algorithm}


\subsubsection{Adaptive time scheduler}
As shown in~\eqref{eq:fm_3}, we typically adopt a uniform time interval 
$\tau$ when solving the ODE for inference. However, the learned model dynamics 
are not uniform in time: in certain regions, the velocity field changes rapidly 
and requires a finer temporal resolution, while in other regions, it evolves 
smoothly and can be handled with coarse steps. To address this, we propose an 
adaptive time scheduler (ATS), which dynamically allocates ODE sampling 
steps according to the temporal complexity of the learned dynamics. It assigns 
more steps where the model is difficult to predict and fewer steps where it is 
smooth.

To construct the ATS, we connect the time scheduler to the training loss in~\eqref{eq:loss_1}. Let the per-timestep MSE loss be denoted as a 
time-dependent function $\mathcal{L}_{\text{MSE}}(t)$. We convert this into a 
difficulty density, which reads
\begin{equation}
\omega(t_i)
=
\left(\mathcal{L}_{\text{MSE}}(t_i)+\psi\right)^{\beta},
\qquad
\beta>0,\ \psi\ge 0,
\label{eq:time_sche_density}
\end{equation}
where $\{t_i\}_{i=0}^{K}$ is a uniformly sampled training grid on $[0,1]$. 
The hyperparameter $\beta$ controls the sharpness of the density and $\psi$ 
prevents degeneracy.

From this density, we form a discrete cumulative distribution function (CDF), given by
\begin{equation}
F_i
=
\frac{\sum_{j=0}^{i}\omega(t_j)}
     {\sum_{j=0}^{K}\omega(t_j)},
\qquad
i = 0,\dots,K.
\label{eq:time_sche_cdf}
\end{equation}
It satisfies $F_0=0$ and $F_K=1$. This CDF is monotonously increasing and 
reflects the relative difficulty of different time intervals.

Given a target sampling budget of $S$ inference steps, we define uniformly 
spaced mass levels
\[
o_s = \frac{s}{S},
\qquad
s=0,\dots,S.
\]
For each $o_s$, we find the interval $[F_{i-1},F_i]$ such that 
$F_{i-1}\le o_s < F_i$ and compute the corresponding adaptive time using 
a piecewise-linear inverse transform sampling. That is, 
\begin{equation}
t_s^{*}
=
t_{i-1}
+
\frac{o_s - F_{i-1}}
     {F_i - F_{i-1}}
\left(t_i - t_{i-1}\right).
\label{eq:time_sche_inverse}
\end{equation}
The resulting sequence $0 = t_0^{*} < t_1^{*} < \dots < t_S^{*} = 1$, constitutes our ATS.

To compute $\mathcal{L}_{\text{MSE}}(t_i)$ in practice, we freeze the 
pretrained model and evaluate it over the training dataset on a fixed uniform 
time grid of $K=50$ points (empirically chosen). We then construct the density 
$\omega(t_i)$, build the CDF in~\eqref{eq:time_sche_cdf}, and finally 
obtain the inference-time schedule $\{t_s^{*}\}_{s=0}^{S}$ using~\eqref{eq:time_sche_inverse}. This reparameterization allocates more ODE 
steps to regions with high training difficulty and fewer steps to smooth 
regions, improving reconstruction fidelity without additional training.

\begin{algorithm}[t]
\caption{Sampling with Adaptive Time Scheduler}\label{ALG:ats}
\begin{algorithmic}
\Require $\boldsymbol{x}_0 \sim \mathcal X_\text{sparse}$
\State $\Tilde{\boldsymbol{x}}_0 = \text{mid}(\boldsymbol{x}_0, \eta)$

\State Compute per-timestep loss $\mathcal{L}_{\text{MSE}}(t_i)$ on the training grid $\{t_i\}_{i=0}^{K}$
\State Compute difficulty weights $\omega_i$ using~\eqref{eq:time_sche_density}
\State Build the discrete CDF $F_i$ using~\eqref{eq:time_sche_cdf}
\State Define adaptive time schedule $\{t_i^{*}\}_{i=0}^{K}$ via inverse transform sampling using~\eqref{eq:time_sche_inverse}
\For{$k = 0$ \textbf{to} $K-1$}
    \State $\delta = t_{k+1} - t_k$
    \State $\kappa = \text{CurvatureEstimate}(\boldsymbol{x}_{t_k})$ \Comment{per-point curvature}
    \State $\boldsymbol{w} = 1 + \alpha_\mathrm{cur} \kappa$ \Comment{curvature-based weights}
    \State $\boldsymbol{x}_{t_{k+1}} = \boldsymbol{x}_{t_k} + \delta (\boldsymbol{w} \odot \nu_\theta(\boldsymbol{x}_{t_k}, t_k))$
\EndFor
\State --------------------------\textbf{Post processing}--------------------------
\State $\boldsymbol{x}_{k\mathrm{NN}}=k\mathrm{NN}(\boldsymbol{x}_{t_{k+1}}, \Tilde{\boldsymbol{x}}_0)$ \Comment{$k$NN search}
\State Compute manifold loss $\mathcal{L}_{\text{manifold}} = \|\boldsymbol{x}_{k\mathrm{NN}} - \Tilde{\boldsymbol{x}}_0\|_2$
\State Update $\boldsymbol{x}_{t_{k+1}} \leftarrow \boldsymbol{x}_{t_{k+1}} - \alpha \nabla_{\boldsymbol{x}_{t_{k+1}}} \mathcal{L}_{\text{manifold}}$
\end{algorithmic}
\end{algorithm}

\begin{figure*}[t]
	\centering
		\centerline{\includegraphics[width=0.8\textwidth]{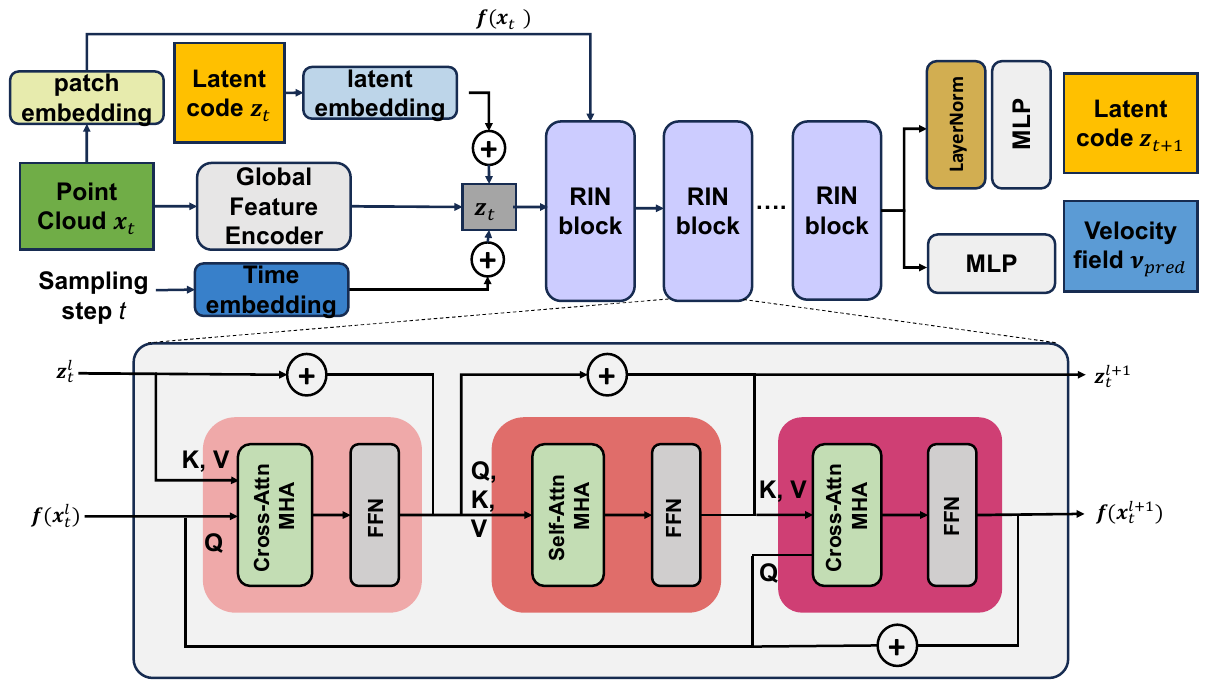}}
		\caption{\textbf{Overview of the proposed iterative point cloud upsampling network.} The network takes the current sampled point cloud $\mathbf{x}_t$, sampling step $t$, and previous latent code $z$ as input. A point feature encoder extracts features $f(\mathbf{x}_t)$. The latent interface $z_t$ is initialized by conditioning on time and global embedding. The core processing consists of stacked RIN blocks. The network outputs the estimated velocity field $\nu_\theta$ and the updated latent code $z_{t+1}$ for the next iteration. }
		
		\label{fig:network}
\end{figure*}

\subsubsection{Testing time optimization and post processing}
We also introduce additional manifold constraints into the sampling process to ensure surface smoothness after point cloud upsampling. Firstly, we estimate curvature on the updated point cloud and learn a weighted velocity for updating. Let the model-predicted point cloud be denoted by $\boldsymbol{x}_t=\{\boldsymbol{a}_{i=1}^N\}$. Then, we estimate the curvature score $\kappa$ by computing 
%
\begin{equation}
\begin{aligned}
& \mathbf{C}_i = \frac{1}{k} \sum_{\boldsymbol{a}_j \in \mathcal{N}_k(\boldsymbol{a}_i)}
(\boldsymbol{a}_j - \bar{\boldsymbol{a}}_i)(\boldsymbol{a}_j - \bar{\boldsymbol{a}}_i)^\top, \\
& \mathbf{C}_i \boldsymbol{v}_j = \lambda_j \boldsymbol{v}_j,
\qquad \text{with} \qquad
\lambda_1 \leq \lambda_2 \leq \lambda_3, \\
& \kappa_i = \frac{\lambda_1}{\lambda_1 + \lambda_2 + \lambda_3},
\label{eq:curve}
\end{aligned}
\end{equation}
where \( \mathcal{N}_k(\boldsymbol{a}_i) \) denotes the \( k \)-nearest neighbors (\( k \)NN) of point \( \boldsymbol{a}_i \), and
\( \bar{\boldsymbol{a}}_i = \frac{1}{k} \sum_{\boldsymbol{a}_j \in \mathcal{N}_k(\boldsymbol{a}_i)} \boldsymbol{a}_j \)
is the local mean of these neighbors.
We compute the local covariance matrix \( \mathbf{C}_i \) and apply eigen decomposition to obtain the principal eigenvalues
\( \lambda_{\{1,2,3\}} \).
The curvature score $\kappa$ is now used to weight the estimated velocity $\nu_\theta(\boldsymbol{x}_{t_k}, t_k)$ by a small learning rate $\alpha_\mathrm{cur}$. Further details are presented in Algorithm~\ref{ALG:ats}, where the curvature-based weights $\omega$ is multiplied with the estimated velocity for point cloud updating ($\odot$ denotes the element-wise multiplication).

In addition, given the $t-$th upsampled point cloud $\boldsymbol{x}_t$, we use $k$NN again to find neighborhoods in the original sparse input $\Tilde{\boldsymbol{x}}_0$, and we define the manifold constraint as the distance between them. We calculate the distance gradient with respect to the upsampled point cloud, then obtain the correct direction that moves the upsampled points closer to the underlying surface manifold. We update the point cloud with a small learning rate $\alpha$ (see post-processing in Algorithm~\ref{ALG:ats}).

\subsection{Latent-state Recurrent Velocity Estimation}
\label{subsec:rin_3d}
Standard flow matching models typically estimate the velocity field $\nu_\theta$ as a stateless function of the current geometry $\boldsymbol{x}_t$ and time $t$. However, point cloud generation is an inherently evolving process; relying solely on the instantaneous state $\boldsymbol{x}_t$ forces the network to re-infer global structure from scratch at every step, often leading to temporal inconsistencies. To address this, we propose a state-dependent estimator that explicitly maintains a global context. We extend the formulation to have $\nu$ additionally depend on $\boldsymbol{z}_t$, where the network predicts both the velocity and the updated latent state for the next step, i.e., 
\begin{equation}
    [\nu_{\text{pred}}, \boldsymbol{z}_{t+1}] = \nu_\theta(\boldsymbol{x}_t, \boldsymbol{z}_t, t).
    \label{eq:inference_step}
\end{equation}
Here, $\boldsymbol{z}_t$ acts as an evolving memory and its initialization $\boldsymbol{z}_{\text{init}}$ comprises sinusoidal time embeddings and global features. This mechanism ensures that the velocity prediction is guided by the accumulated geometric history. To implement this state-dependent function, we adopt a recurrent interface network (RIN)~\cite{rin} architecture, which utilizes a set of learnable latent tokens to represent $\boldsymbol{z}$, and interacts with point features $f(\boldsymbol{x}_t^l)$ via a stacked Read-Compute-Write attention mechanism (where $l$ denotes the $l$-th layer feature maps), which reads
\begin{align}
& \text{Read:} && \boldsymbol{z}_t^{l+1} = \boldsymbol{z}_t^l + \operatorname{MHA}(\boldsymbol{z}_t, f(\boldsymbol{x}_t^l)) \nonumber \\
&&& \boldsymbol{z}_t^{l+1} = \boldsymbol{z}_t^l + \operatorname{MLP}(\boldsymbol{z}_t^l) \nonumber \\
& \text{Compute:} && \boldsymbol{z}_t^{l+1} = \boldsymbol{z}_t^l + \operatorname{MHA}(\boldsymbol{z}_t^l, \boldsymbol{z}_t^l) \nonumber \\
&&& \boldsymbol{z}_t^{l+1} = \boldsymbol{z}_t^l + \operatorname{MLP}(\boldsymbol{z}_t^l) \nonumber \\
& \text{Write:} && f(\boldsymbol{x}_t^{l+1}) = f(\boldsymbol{x}_t^l) + \operatorname{MHA}(f(\boldsymbol{x}_t^l), \boldsymbol{z}_t^l) \nonumber \\
&&& f(\boldsymbol{x}_t^{l+1}) = f(\boldsymbol{x}_t^l) + \operatorname{MLP}(f(\boldsymbol{x}_t^l)),
\label{eq:rin}
\end{align}
where $\operatorname{MHA}$ stands for multi-head attention. This structure allows the network to aggregate sparse local information into a dense global representation (Read), process it abstractly (Compute), and redistribute the refined context back to the points (Write).

\textbf{Training with two-pass estimation.} 
A challenge arises during training because time steps $t$ are sampled independently, making the sequential history $\boldsymbol{z}_{t-1}$ unavailable. To simulate recurrence, we employ a two-pass strategy. In the first pass, we estimate a proxy latent state $\tilde{\boldsymbol{z}}_t$ using a null initialization. That is,
\begin{equation}
    \tilde{v}_\mathrm{pred}, \tilde{\boldsymbol{z}}_t = \nu_\theta(\boldsymbol{x}_t, \mathbf{0}, t).
\end{equation}
In the second pass, this proxy latent,  acting as the ``remembered'' context, is used to condition the network for the final prediction, which reads
\begin{equation}
    \nu_{\text{pred}}, \boldsymbol{z}_{t+1} = \nu_\theta(\boldsymbol{x}_t, \operatorname{sg}(\boldsymbol{z}_t), t),
\end{equation}
where $\operatorname{sg}(\cdot)$ denotes the stop-gradient operation. This effectively trains the model to utilize its own latent predictions, bridging the gap between parallel training and sequential inference.

\section{Experiments}
\label{sec:exps}
\begin{table*}[t]
\caption{\textbf{Point cloud upsampling comparison with state-of-the-art methods.} We report the numbers of the Chamfer distance (CD, scaled by $10^{4}$), the Hausdorff distance (HD, scaled by $10^{3}$), point-to-surface (P2F, scaled by $10^{3}$), and the Jensen-Shannon divergence (JSD without scaling) on the PUGAN and PU1K datasets. Red indicates the best results, and blue indicates the second-best results. Note that the comparisons are conducted on arbitrarily downsampled sparse point clouds; therefore, the results of different methods are not directly comparable to those reported in the literature.}
\centering
\renewcommand\arraystretch{1.3}
\resizebox{\linewidth}{!}{
\begin{tabular}{c|c|ccccccccccc}
\toprule
Dataset & \begin{tabular}[c]{@{}c@{}}upsampling \\ factor\end{tabular} & Metrics & PUBP & PUGCN & RepKPU & PUGAN & PUDM & Grad-PU & PUFM & SAPCU & PDANS$^{\dagger}$ & Ours \\ \midrule
\multirow{8}{*}{PUGAN} & \multirow{4}{*}{4} & CD & 1.649 & 2.774 & 1.067 & 1.541 & 1.221 & 1.132 & \textcolor{blue}{1.049} & 1.522  & 1.201 & \textcolor{red}{0.980} \\
 &  & HD & 1.476 & 3.831 & 1.139 & 1.391 & 1.174 & 1.186 & \textcolor{blue}{0.876} & 1.471  & 1.165 & \textcolor{red}{0.747} \\
 &  & P2F & 5.997 & 9.508 & 1.974 & 5.420 & 3.132 & 1.957 & \textcolor{blue}{1.864} & 3.441  & 2.897 & \textcolor{red}{1.301} \\
  &  & JSD & 0.208 & 0.205 & 0.196 & 0.196 & 0.147 & \textcolor{blue}{0.111} & 0.121 & 0.135  & 0.148 & \textcolor{red}{0.098} \\
 & \multirow{4}{*}{16} & CD & 0.982 & 1.102 & 0.384 & 0.869 & 0.533 & 0.415 & \textcolor{blue}{0.353} &  0.717 & 0.498 & \textcolor{red}{0.286} \\
 &  & HD & 2.071 & 1.785 & 1.245 & 1.746 & 1.185 & 1.142 & \textcolor{blue}{0.844} & 5.371 & 1.175 & \textcolor{red}{0.687} \\
 &  & P2F & 7.496 & 7.125 & 2.151 & 6.757 & 3.589 & 2.185 & \textcolor{blue}{2.103} & 3.256 & 3.226 & \textcolor{red}{1.412} \\
 &  & JSD & 0.149 & 0.250 & 0.077 & 0.143 & 0.108 & 0.056 & 0.074 & \textcolor{blue}{0.070} & 0.089 & \textcolor{red}{0.040} \\
 \midrule
\multirow{8}{*}{PU1K} & \multirow{4}{*}{4} & CD & 0.694 & 1.241 & 0.566 & 0.682 & 0.706 & 0.626 & \textcolor{blue}{0.545} & 0.605 & 0.669 & \textcolor{red}{0.491}  \\
 &  & HD & 0.593 & 1.504 & 0.619 & 0.632 & 0.605 & 0.583 & \textcolor{blue}{0.556} & 0.587 & 0.600 & \textcolor{red}{0.432} \\
 &  & P2F & 2.206 & 5.115 & 1.464 & 2.792 & 2.891 & 1.510 & 1.770 & \textcolor{blue}{1.655} & 2.884 & \textcolor{red}{1.243} \\
 &  & JSD & 0.286 & 0.359 & 0.265 & 0.315 & 0.351 & \textcolor{blue}{0.262} & 0.304 & 0.361 & 0.352 & \textcolor{red}{0.256} \\
 & \multirow{4}{*}{16} & CD & 0.343 & 2.393 & 0.290 & 0.361 & 0.421 & 0.316 & \textcolor{blue}{0.220} & 0.379 & 0.415 & \textcolor{red}{0.176} \\
 &  & HD & 0.712 & 4.214 & 0.592 & 0.773 & 0.602 & \textcolor{blue}{0.552} & 0.578 & 2.006 & 0.589 & \textcolor{red}{0.429} \\
 &  & P2F & 2.958 & 9.410 & 1.821 & 3.538 & 3.370 & \textcolor{blue}{1.890} & 1.983 & 2.400 & 3.320 & \textcolor{red}{1.375}\\
 &  & JSD & 0.282 & 0.362 & 0.229 & 0.225 & 0.287 & \textcolor{blue}{0.211} & 0.228 & 0.262 & 0.285 & \textcolor{red}{0.135} \\ 
 \bottomrule
\end{tabular}%
}
\label{tab:pc_sota}
\end{table*}

\begin{figure*}[t]
	\centering
		\centerline{\includegraphics[width=\textwidth]{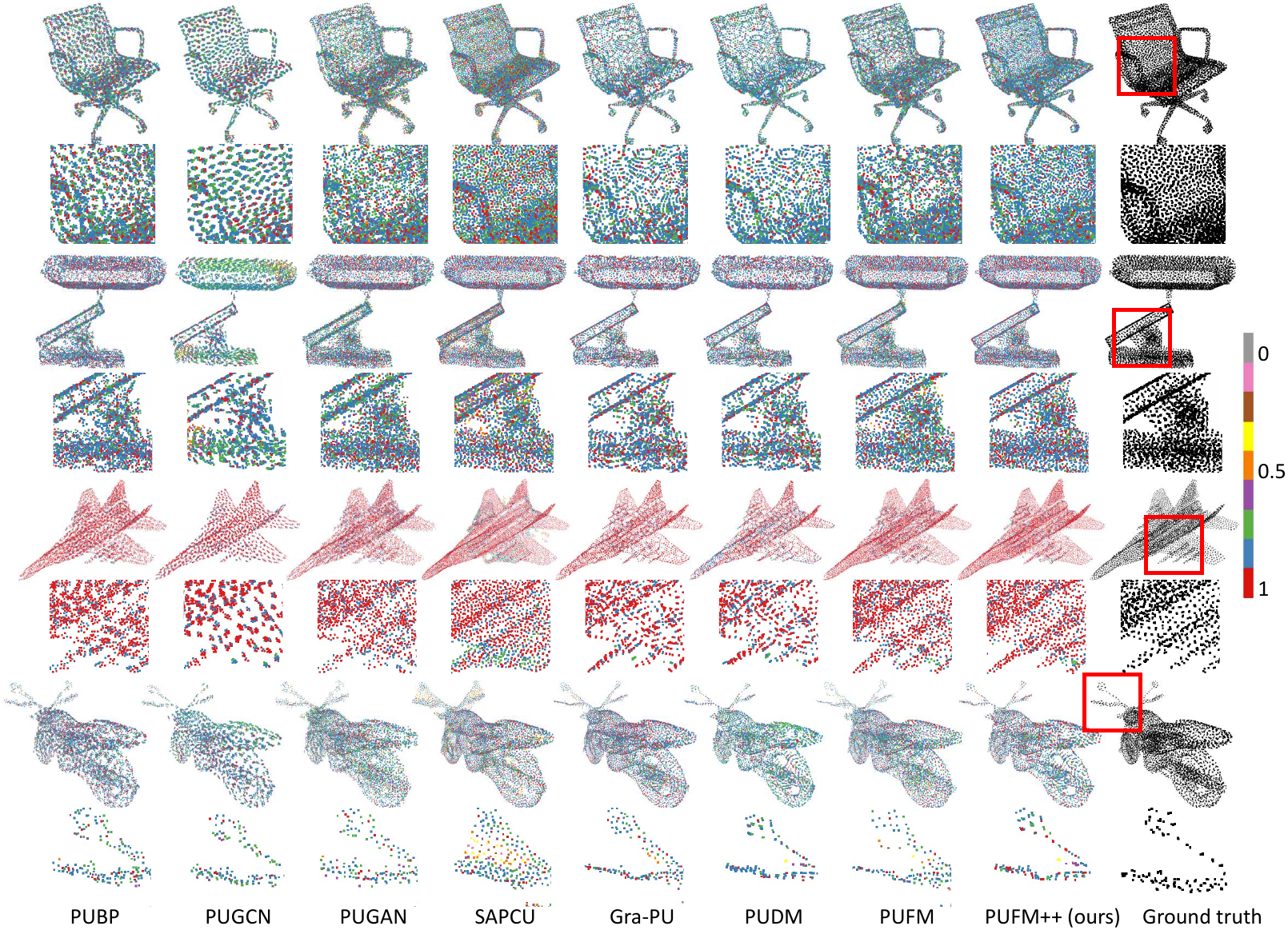}}
		\caption{\textbf{Visual comparison of different methods on 4-times upsampling.} We apply different methods to four examples from the PU1K dataset. Our method generates evenly distributed points and sharp edges, while others exhibit obvious holes and noisy surfaces (see the enlarged region in red boxes).
		}
		\label{fig:sota}
\end{figure*}

\subsection{Experimental Details}
\textbf{Dataset.} We use two publicly available datasets for training and evaluation, {PUGAN}~\cite{pugan} and {PU1K}~\cite{pu-gcn}. We follow the same procedure as in ~\cite{mpu} to extract paired sparse and dense point clouds. Given the 3D meshes, we first use Poisson disk sampling to generate dense point clouds, then use farthest point sampling (FPS) to select centroids, and finally use $k$NN to form dense patches for all centroids. Each dense patch contains 1024 points. Then we randomly sample 256 points from each dense patch to obtain the corresponding sparse patch. For testing, we obtain 27 and 127 point clouds from PUGAN and PU1K, respectively. Moreover, we also use three real-world point clouds for further analysis using the datasets {ScanNet}~\cite{scannet}, {KITTI}~\cite{kitti}, and {ABC}~\cite{abc}.

\noindent \textbf{Evaluation metrics.} To measure point cloud upsampling quality, we use the Chamfer distance ({CD}), the Hausdorff distance ({HD}), and point-to-surface ({P2F}) as the evaluation metrics in our experiments. Additionally, we also use the Jensen-Shannon divergence (JSD) score to measure the point cloud distribution. We divide the 3D space into voxels using a $k$-d tree, then measure the average KL divergence between pairs of voxels. When JSD is close to zero, it indicates that the two point clouds are (almost) identical. We are also interested in mesh reconstruction from upsampled point clouds for downstream applications. Hence, we use normal consistency ({NC}), i.e., the mean absolute cosine of normals in one mesh and normals at nearest neighbors in the other mesh. The area-length ratio ({ALR})~\cite{tetsphere} measures the shape quality of individual triangles within a mesh. This ratio attains its maximum value of 1 for equilateral triangles, indicating optimal shape quality. The manifoldness rate ({MR})~\cite{tetsphere} quantifies the proportion of edges in a mesh that satisfy manifold conditions. An edge is considered a manifold if exactly two faces share it. A rate approaching 1 indicates a well-structured mesh suitable for operations like Boolean computations and 3D printing. Conversely, a lower rate highlights the presence of non-manifold edges, which may necessitate mesh repair or refinement. 

\noindent \textbf{Training settings.} We modify the RIN~\cite{rin} architecture and use it as our velocity field model $\nu_\theta$. As summarized in Algorithm \textcolor{red}{1}, the training process comprises two stages. For stage 1 of pre-aligned flow matching optimization, we mainly follow the setting in PUFM~\cite{pufm} and use a batch size of 32 and a learning rate of $10^{-4}$. We train the model for 100 epochs with the Adam optimizer~\cite{adam}. The average losses for different sampling steps are used for CDF computation (Equation~\ref{eq:time_sche_cdf}) and saved for the inference stage. For state 2 of endpoint flow matching refinement, we fine-tune the model using a learning rate of $10^{-5}$ for 50 epochs. We define the noise scale $\sigma=0.02$ for the additional Gaussian noise. For data augmentation, we randomly apply rotation and shifting to the sparse and dense point clouds. 

\noindent \textbf{Inference settings.} For the inference, we first apply FPS and kNN to the target sparse point clouds and obtain patches as model inputs. Based on Algorithm \textcolor{red}{2}, we use the CDF and total number of inference steps to calculate the adaptive time steps for ODE sampling. In the last step, it is optional to apply manifold optimization for further improvement. We define the updating factor $\alpha=0.01$ based on the empirical results. Finally, we assemble all the upsampled patches to form the final dense point clouds.

\begin{figure*}[t]
	\centering
		\centerline{\includegraphics[width=\textwidth]{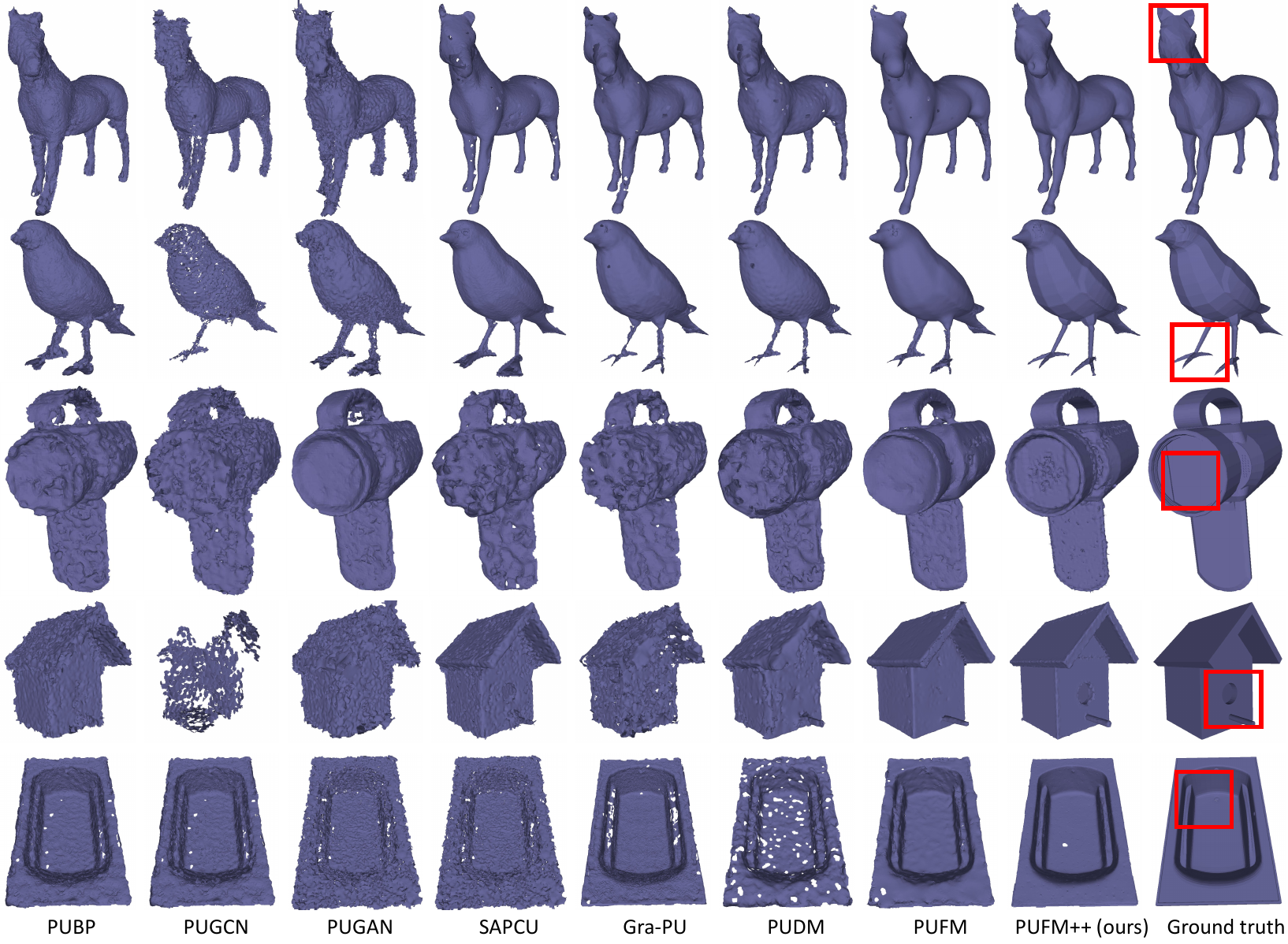}}
		\caption{\textbf{Visual comparison of different methods on 16-times upsampling for mesh reconstruction.} We apply different methods to five examples from the PUGAN and PU1K dataset. Then we use the Marching Cubes algorithm to reconstruct meshes from the upsampled point clouds for qualitative comparison.
		}
		\label{fig:sota_mesh}
\end{figure*}

\begin{table*}[t]
\caption{\textbf{Mesh reconstruction comparison with state-of-the-art methods.} We show the area-length ratio (ALR), the manifoldness rate (MR), and normal consistency (NC) on the PU1K datasets. Red indicates the best results, and blue indicates the second-best results.}
\centering
\renewcommand\arraystretch{1.3}
\resizebox{\linewidth}{!}{
\begin{tabular}{c|c|ccccccccccc}
\toprule
Dataset & \begin{tabular}[c]{@{}c@{}}upsampling \\ factor\end{tabular} & Metrics & PUBP & PUGCN & RepKPU & PUGAN & PUDM & Grad-PU & PUFM & SAPCU & PDANS$^{\dagger}$ & Ours \\ \midrule
\multirow{6}{*}{PU1K} & \multirow{3}{*}{4} & ALR & 0.231 & 0.202 & \textcolor{red}{0.251} & 0.237 & 0.239 & 0.229 & 0.243 & 0.243 & 0.240 & \textcolor{blue}{0.250} \\
 &  & MR & 0.990 & 0.989 & 0.990 & \textcolor{blue}{0.991} & 0.989 & 0.990 & \textcolor{blue}{0.991} & \textcolor{blue}{0.991} & 0.988 & \textcolor{red}{0.992} \\
 &  & NC & 0.946 & 0.932 & \textcolor{blue}{0.949} & 0.944 & 0.930 & 0.939 & \textcolor{blue}{0.949} & 0.941 & 0.931 & \textcolor{red}{0.957} \\
 & \multirow{3}{*}{16} & ALR & 0.256 & 0.216 & 0.240 & 0.246 & 0.242 & 0.235 & 0.237 & \textcolor{blue}{0.246} & 0.243 & \textcolor{red}{0.267} \\
 &  & MR & \textcolor{blue}{0.991} & 0.989 & \textcolor{blue}{0.991} & \textcolor{blue}{0.991} & 0.988 & 0.989 & \textcolor{blue}{0.991} & \textcolor{blue}{0.991} & 0.988 & \textcolor{red}{0.993} \\
 &  & NC & 0.922 & 0.846 & 0.932 & 0.909 & 0.886 & 0.896 & 0.921 & \textcolor{blue}{0.929} & 0.891 &  \textcolor{red}{0.950} \\ \bottomrule
\end{tabular}%
}
\label{tab:mesh_sota}
\end{table*}

\subsection{Comparison to the State of the Art}
Here, we evaluate PUFM++ on two datasets, PUGAN and PU1K, and compare it to the state-of-the-art approaches PUBP~\cite{dualbp}, PUGCN~\cite{pu-gcn}, RepKPU~\cite{repkpu}, PUGAN~\cite{pugan}, PUDM~\cite{pudm}, Grad-PU~\cite{grad-pu}, PUFM~\cite{pufm}, SAPCU~\cite{sapcu} and PDANS~\cite{pdans}. Note that the authors of PDANS do not release their pretrained models, so we reimplemented their model based on the official code and reported the results as PDANS$^{\dagger}$. Table~\ref{tab:pc_sota} reports the results (including 4-times and 16-times upsampling) on the point cloud-based metrics CD, HD, P2F, and JSD. We can see that our approach achieves notable margins on all metrics, especially on P2F. Specifically, from CD and HD scores, we can see that our method outperforms PUFM by +0.07 and +0.13 on 4-times upsampling, and +0.05 and +0.12 on 16-times upsampling. These margins are even larger than those of other methods. For the P2F score, the most competitive method is Grad-PU. However, our method still achieves the best results, with improvements of $0.3$--$0.7$ in two upsampling scenarios. We also add the JSD score as a distribution similarity measure, and we find that ours achieves the lowest value among all approaches. Specifically when we compare it with PUFM, PUFM++ reduces the JSD score by $20$--$45\%$.

Figure~\ref{fig:sota} shows the visual comparison of 4-times point cloud upsampling. To highlight the nuanced differences, we first calculate the point-to-point distance between the true data and the model prediction, then normalize the distances and encode them into different colors. We can conclude that 1) from the point distribution, our method has smooth and uniform distributions even when the input point clouds contain hole-like uneven distributions; cf., e.g., the motorcycle; 2) from the general color distribution on the point clouds, see, e.g., the chair in the first row, our approach achieves the lowest errors out of all approaches; 3) compared to PUFM, our approach can preserve sharp 3D shapes, cf., e.g., the plane, and remove outliers, as for the motorcycle; 4) SAPCU tends to oversmooth the regions without considering the sharp curvatures, see, e.g., the lamp, and PUDM preserves the uneven distributions which causes the unsmooth surface, as for the chair and the motorcycle.

Besides, we are also interested in the mesh reconstruction from the upsampled point clouds. To generate the meshes, we first compute the normals from the point cloud, then generate the meshes using the Marching Cube method. We report the mesh comparison in Table~\ref{tab:mesh_sota} across 4-times and 16-times upsampling. For normal estimation (NC), our approach outperforms all other methods by significant margins, indicating its surface smoothness. For ALR, PUFM++ is the second-best one behind the RepKPU approach. For MR, PUFM++ achieves performance equal to or better than others. Figure~\ref{fig:sota_mesh} shows the visual comparison on mesh reconstruction. We can see that ours can produce meshes with smoother surfaces and fewer holes. On the contrary, other methods produce noisy, incomplete meshes (see the regions marked by the red boxes). For example, our method can estimate the fine details of the horse (first row) and bird (second row) around the legs. It can also produce smoother surfaces for the camera (third row), house (fourth row), and the plate (fifth row).

\subsection{Ablation of PUFM++}
In this section, we provide ablations of PUFM++. Specifically, we ablate the training schemes~\ref{Training schemes} and inference schemes~\ref{Inference schemes}.

\subsubsection{Training schemes}
\label{Training schemes}
For the training stage, we propose two significant improvements: 1) two-stage flow matching optimization with pre-alignment, and 2) replacing the PointNet++ with the Recurrent Interface Network (RIN).

\noindent \textbf{Two-stage flow matching optimization.} As introduced in Section~\ref{sec:method}, one of our key claims is the two-stage training approach. In stage 1, we perform pre-alignment to improve linear interpolation between sparse and dense point clouds. The pre-alignment is an approximation of the EMD, applied only during training with very little computational overhead. In the inference, we can skip pre-alignment and directly refine sparse point clouds to obtain dense output. In the second stage, we have the endpoint flow-matching refinement, which further improves upsampling quality. To make a comparison, we report the ablation results in Table~\ref{tab:ablation_train}.

\begin{table}[t]
\caption{\textbf{Ablation on training strategies.} CD, HD, and P2F scores on PUGAN and PU1K datasets.}

\centering
\renewcommand\arraystretch{1.7}
\resizebox{\linewidth}{!}{
\begin{tabular}{cccc|cccccc}
\toprule
\multicolumn{4}{c|}{Training strategy} & \multicolumn{3}{c}{PUGAN} & \multicolumn{3}{c}{PU1K} \\
NoAlign & CD & EMD* & S2-Refine & CD & HD & P2F & CD & HD & P2F \\ \midrule
\checkmark &  &  &  & 4.989 & 4.335 & 2.120 & 3.001 & 4.189 & 8.44 \\
 & \checkmark &  &  & 1.135 & 1.200 & 1.895 & 0.648 & 0.756 & 1.766 \\
 &  & \checkmark &  & 1.010 & 0.815 & 1.742 & 0.523 & 0.522 & 1.658 \\
 &  &  & \checkmark & 1.012 & 0.801 & 1.754 & 0.526 & 0.516 & 1.716 \\
\cellcolor{mistyrose}{} & \cellcolor{mistyrose}{} & \cellcolor{mistyrose}{\checkmark} & \cellcolor{mistyrose}{\checkmark} & \cellcolor{mistyrose}{0.098} & \cellcolor{mistyrose}{0.747} & \cellcolor{mistyrose}{1.301} & \cellcolor{mistyrose}{0.491} & \cellcolor{mistyrose}{0.432} & \cellcolor{mistyrose}{1.243} \\ \bottomrule
\end{tabular}%
}
\label{tab:ablation_train}

\end{table}

\begin{table}[t]
\caption{\textbf{Ablation on network selection.} CD, HD, and P2F scores on the PUGAN and PU1K datasets.}

\centering
\renewcommand\arraystretch{1.3}
\resizebox{\linewidth}{!}{
\begin{tabular}{c|cccccc|cc}
\toprule
\multirow{2}{*}{Network} & \multicolumn{3}{c}{PUGAN} & \multicolumn{3}{c|}{PU1K} & \multirow{2}{*}{Params} & \multirow{2}{*}{\begin{tabular}[c]{@{}c@{}}Running \\ time (s)\end{tabular}} \\
 & CD & HD & P2F & CD & HD & P2F &  &  \\ \midrule
\begin{tabular}[c]{@{}c@{}}PointNet++\\ (single-stage)\end{tabular} & 1.049 & 0.876 & 1.864 & 0.545 & 0.556 & 1.770 & \multirow{2}{*}{30.36} & \multirow{2}{*}{0.71/5} \\
\begin{tabular}[c]{@{}c@{}}PointNet++\\ (two-stage)\end{tabular} & 1.007 & 0.826 & 1.582 & 0.524 & 0.512 & 1.705 &  &  \\ \midrule
\begin{tabular}[c]{@{}c@{}}Rin\\ (single-stage)\end{tabular} & 1.012 & 0.801 & 1.754 & 0.526 & 0.516 & 1.716 & \multirow{2}{*}{115.26} & \multirow{2}{*}{1.231/6} \\
\begin{tabular}[c]{@{}c@{}}RIN\\ (two-stage)\end{tabular} & 0.098 & 0.747 & 1.301 & 0.491 & 0.432 & 1.243 &  & \\
\bottomrule
\end{tabular}%
}
\label{tab:ablation_network}
\end{table}

\begin{figure}[t]
	\centering
		\centerline{\includegraphics[width=\columnwidth]{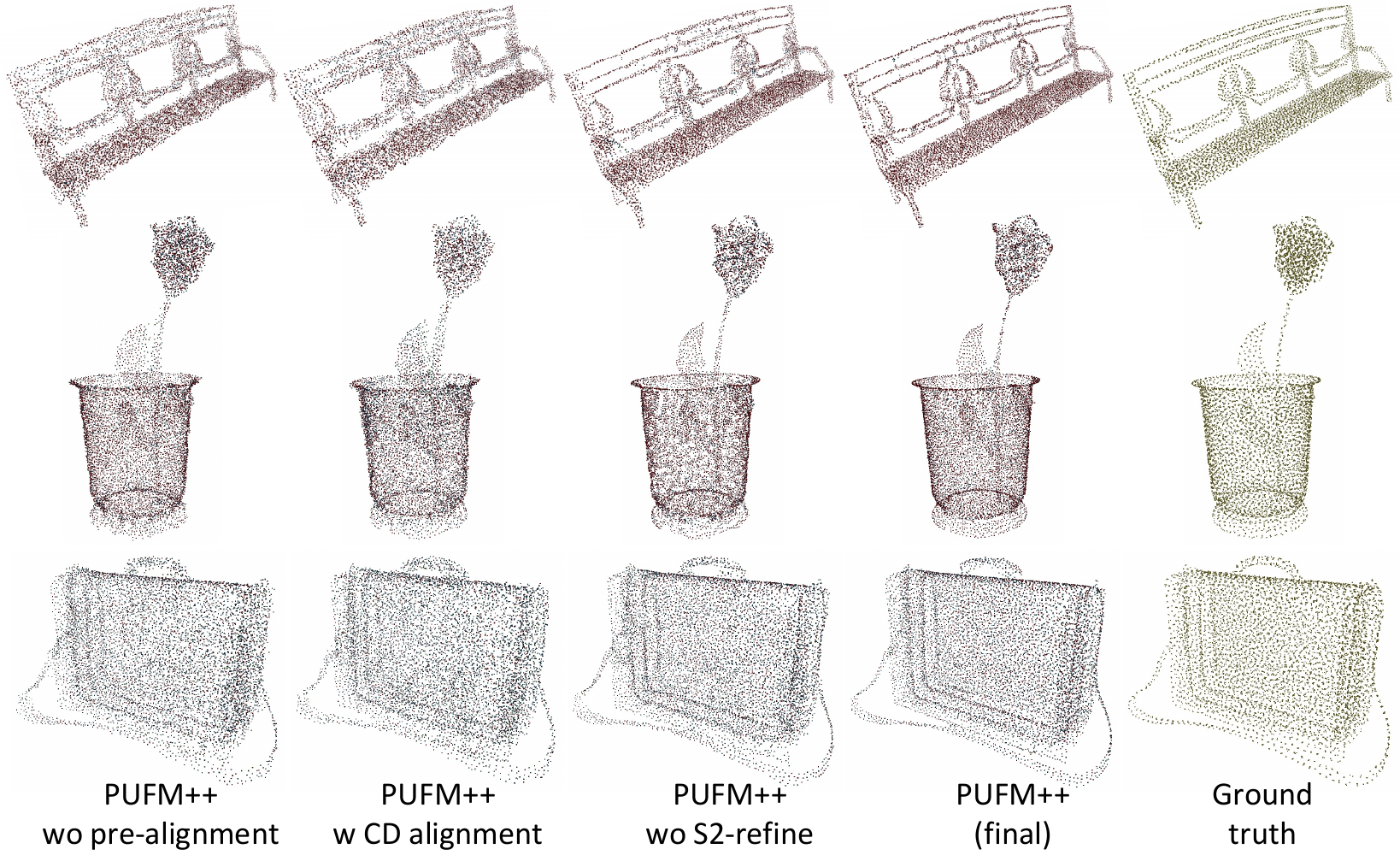}}
		\caption{\textbf{Visual comparison of training with/without pre-alignment and refinement.} The final PUFM++ (4th column) is compared to: PUFM++ without pre-alignment/refinement (1st), with CD pre-alignment only (2nd), and with EMD pre-alignment only (3rd).
		}
		\label{fig:abla_train}
\end{figure}

In Table~\ref{tab:ablation_train}, we compare the full model with or without the pre-alignment and stage two of refinement. For the pre-alignment, we compare the approximated EMD (EMD*) with CD for alignment. We can see that both of them achieve significant improvements over the approach without alignment. Furthermore, using the approximated EMD yields improvements of $0.1-0.4$ in CD and HD scores. Comparing row 3 and row 5, we can also see that using the additional endpoint flow-matching refinement reduces CD loss by 0.06 to 0.12 across two datasets. Visually, we show three examples on 4-times upsampling to compare the final PUFM++ model with or without these key modules. From Figure~\ref{fig:abla_train}, we can see that the model trained without pre-alignment generates noisy and blurry 3D surfaces around the bench (first row) and produces unstable edges (e.g., the flower and the bag in rows 2 and 3, respectively). Incorporating CD pre-alignment slightly reduces noise but does not fully resolve the artifacts. In contrast, comparing the third and fourth columns shows that adding the endpoint refinement stage significantly improves the point distribution along edges, as illustrated by the clearer bench patterns in the first example and the sharper vase contours in the second. 

\noindent \textbf{Recurrent Interface Network for PUFM++.} We propose to replace the PointNet++ with the RIN network in Section 3.3. The main advantage is that it avoids the U-shaped feature extraction and aggregation, and refines the latent representation iteratively before projecting back onto the point space. Meanwhile, the latent self-conditioning stabilizes iterative upsampling, thereby avoiding artifacts such as point clustering or uneven density. In Table~\ref{tab:ablation_network}, we compare upsampling quality between PointNet++ and RIN. We can conclude that 1) the RIN uses 3 times more parameters but achieves 90\% improvements on both PUGAN and PU1K datasets. The running time only increases \roland{to a} 50\% computation overhead (0.14s to 0.21s); 2) both PointNet++ and RIN can boost the performance using two-stage optimization, but using RIN with single-stage optimization achieves similar performance as the PointNet++ with two-stage optimization.

\subsubsection{Inference schemes}
\label{Inference schemes}
For the inference stage, we propose two novel approaches to boost the upsampling quality: 1) ODE inference steps, 2) the adaptive entropic time scheduler, and 3) the manifold constraint.

\begin{figure}[t]
	\centering
		\centerline{\includegraphics[width=\columnwidth]{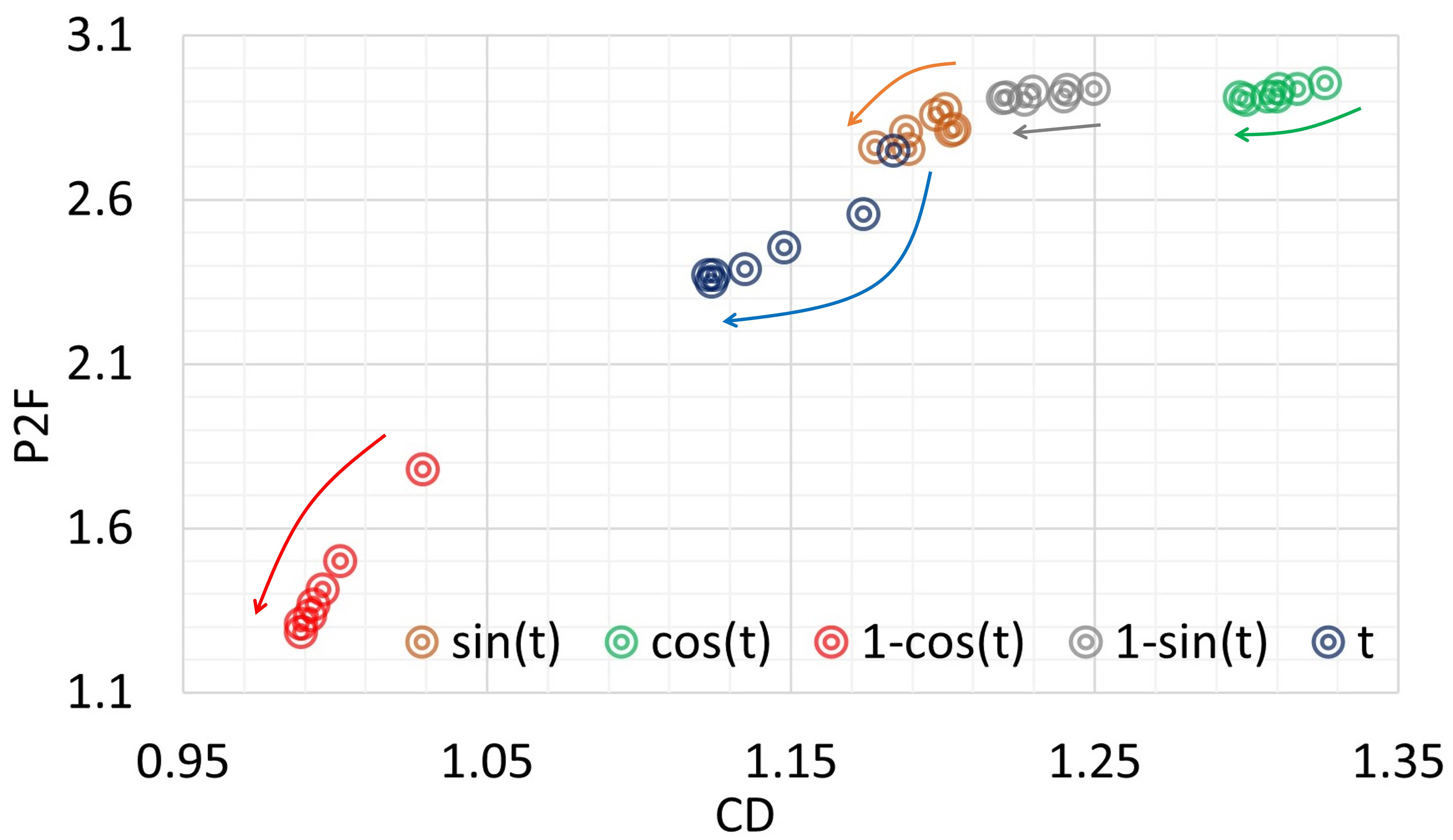}}
		\caption{\textbf{Visual comparison of ODE sampling steps and time scheduler.} We apply different time schedulers to train the proposed model, and then use different discretization steps of the ODE for inference. We plot the 4-times upsampling results on the PUGAN dataset, where the horizontal axis is the CD score and the vertical axis is the P2F distance.
		}
		\label{fig:ode_step}
\end{figure}

\begin{figure*}[t]
	\centering
		\centerline{\includegraphics[width=\textwidth]{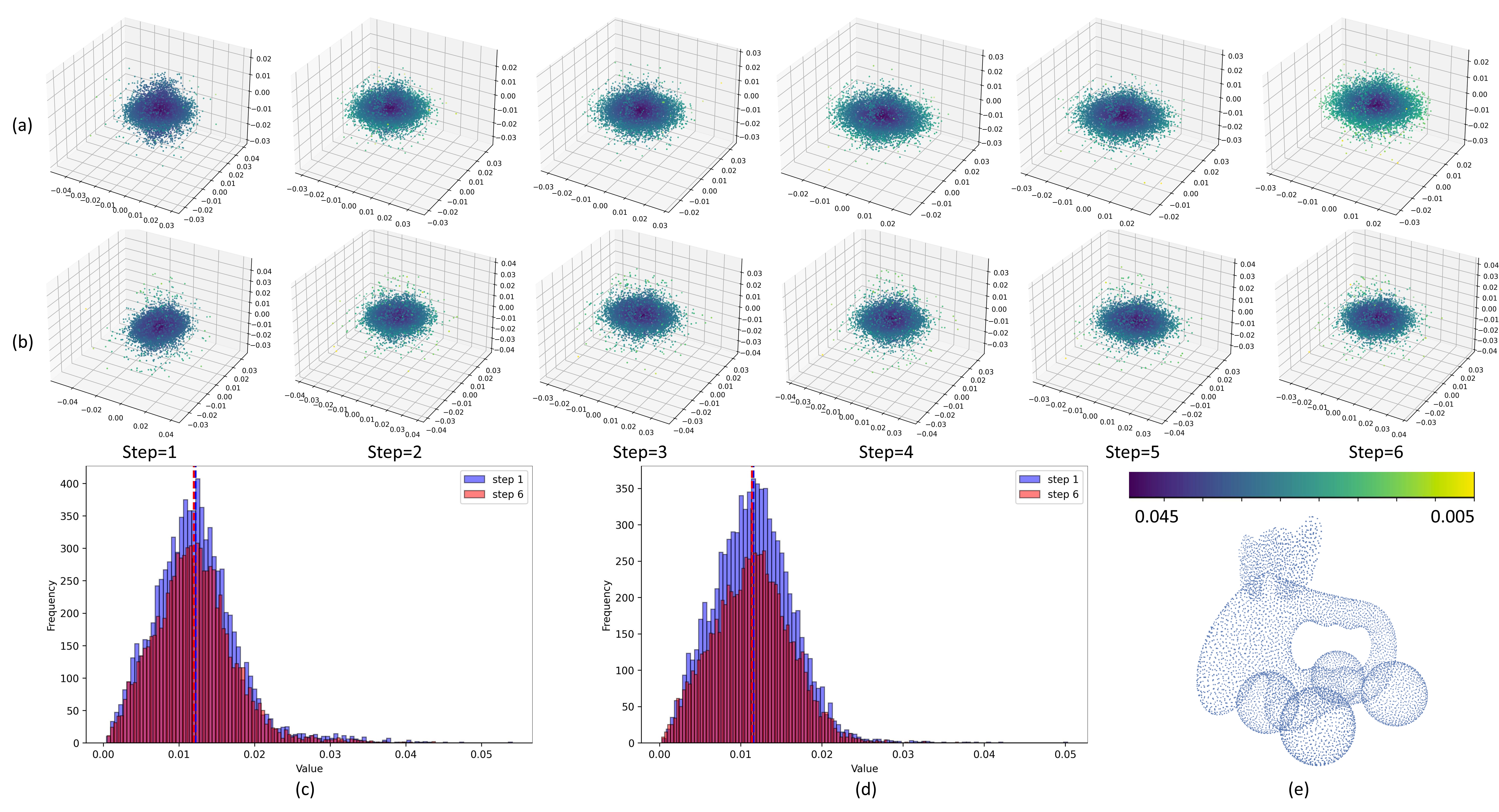}}
		\caption{\textbf{Visualization of entropic time scheduler for inference.} We show the residual scatter map on 4-times upsampling on the point cloud (e) by calculating the coordinate differences using (a) an entropic time scheduler and (b) uniform time scheduler. The darker the color, the bigger the errors. We also show the residual histograms for the two inference schemes in (c) and (d). 
		}
		\label{fig:vis_time_scheduler}
\end{figure*}

\textbf{Inference steps.} The number of inference steps and the time scheduler for linear interpolation used in training affect the point cloud upsampling. In our algorithm (Algorithm \textcolor{red}{1}), we choose $1-\cos(s\pi /2)$ as the time scheduler because we want to encourage the model to learn the target velocity from more challenging points that are closer to the sparse point clouds. During inference, the proposed PUFM++ is similar to other score-based generative models: the more iterations we use, the better the results are. From Figure~\ref{fig:ode_step}, we can see that using $1-\cos(s\pi /2)$ (red circles) lies on the left bottom corner, where it has both a low CD score and a P2F distance. On the other hand, we can see that $t$ (blue), $\sin(s\pi/2)$ (orange), $1-\sin(s\pi/2)$ (green) are all suboptimal. To balance upsampling quality and runtime, we apply 6-step inference. 

\begin{figure}[t]
	\centering
		\centerline{\includegraphics[width=\columnwidth]{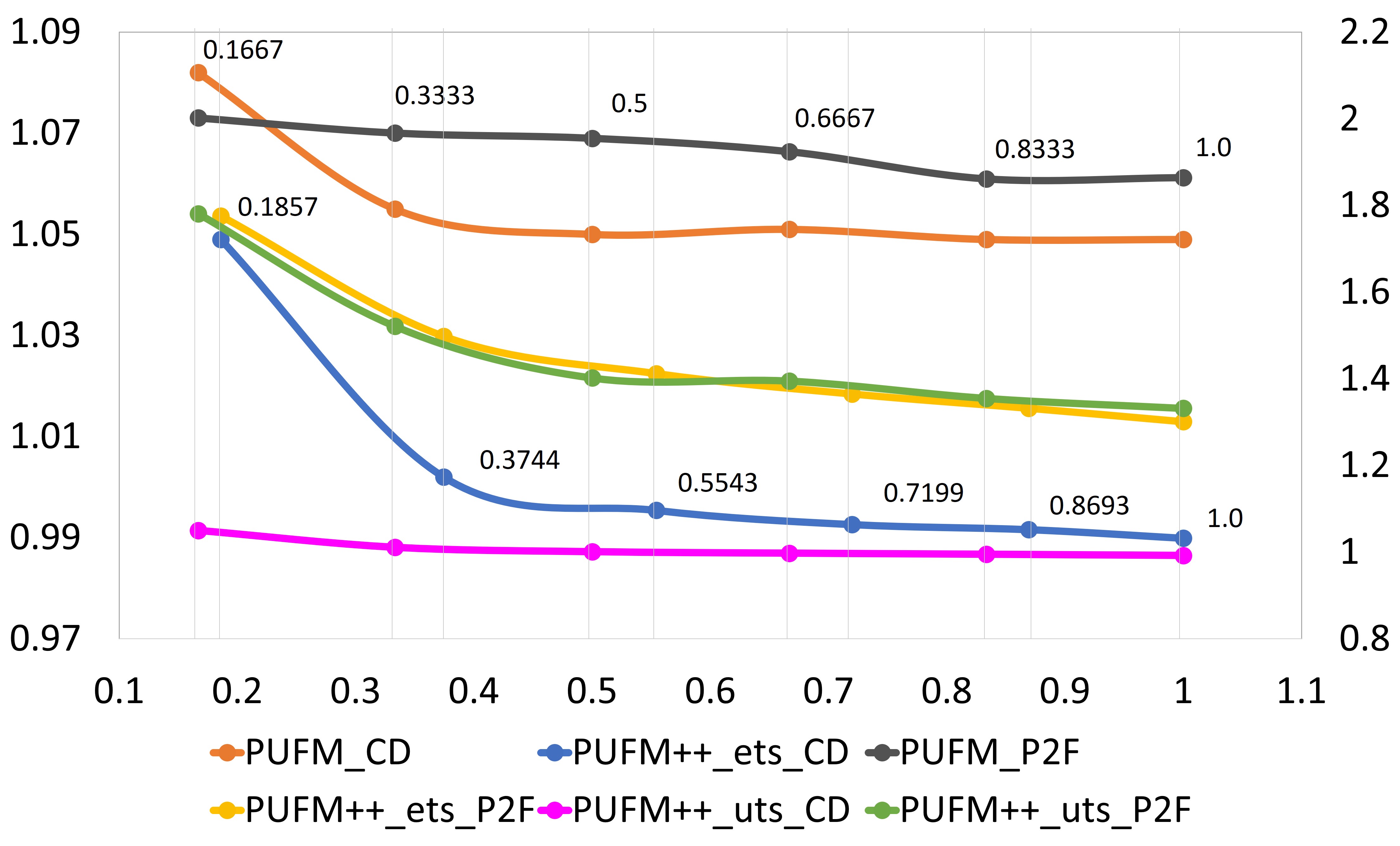}}
		\caption{\textbf{Compare different time schedulers for inference.} We show the per-step inference using the uniform time scheduler and the proposed adaptive time scheduler on PUFM and PUFM++.
		}
		\label{fig:compare_time_scheduler}
\end{figure}

\textbf{Adaptive entropic time scheduler.} As described in Section~\ref{sec:method}, we freeze the pre-trained model and test it on the training dataset, then record the average losses on the training sampling grid $\{t_0, t_1, t_2,\ldots\}$. It is defined as a uniform time sequence sampled from [0, 1] with a total of 50 steps. Then we compute the CDF of the recorded losses $\hat{F}(t)$. During the inference, we compute the time step as in~\ref{eq:time_sche_inverse} for ODE sampling. This approach leverages information from the training process to adjust step sizes, thereby improving inference quality dynamically. In Figure~\ref{fig:compare_time_scheduler}, we show results for PUFM, PUFM++ with a uniform time scheduler ($\text{PUFM++\_uts}$) and the proposed PUFM++ with an entropic time scheduler ($\text{PUFM++\_ets}$). We can see that 1) PUFM, as the baseline model, achieves the worst results at every inference step; 2) compared to $\text{PUFM++\_uts}$, the proposed $\text{PUFM++\_ets}$ allocates larger step sizes in the early ODE iterations and progressively refines them in later stages. This behavior enables the model to focus more on complex refinements near the end of the trajectory. Quantitatively, on $4$-times point cloud upsampling with the PUGAN dataset, our scheduler consistently improves both CD and P2F metrics over uniform time stepping, demonstrating its effectiveness.

We further provide a qualitative example of $4$-times upsampling in Figure~\ref{fig:vis_time_scheduler}. In subfigures (a) and (b), we visualize the coordinate residuals between the upsampled point clouds and the ground-truth target. The residuals are normalized and mapped onto a 3D scatter plot, color-coded by their Euclidean distances. Darker color indicates bigger errors, and tigher clusters indicate smaller errors. We can see from subfigures (a) and (b) that using the entropic time scheduler provides overall smaller errors across all inference steps. Subfigures (c) and (d) show the corresponding histograms of the residuals. Results in (a) and (c) are obtained with the uniform time scheduler, while (b) and (d) correspond to our proposed entropic time scheduler. Compared with uniform scheduling, the entropic scheduler shifts the mean residuals closer to zero. It significantly compresses the overall error distribution (see the reduced maximum values in (b) vs. (a)), indicating more stable and accurate reconstructions.

\begin{figure}[t]
	\centering
		\centerline{\includegraphics[width=\columnwidth]{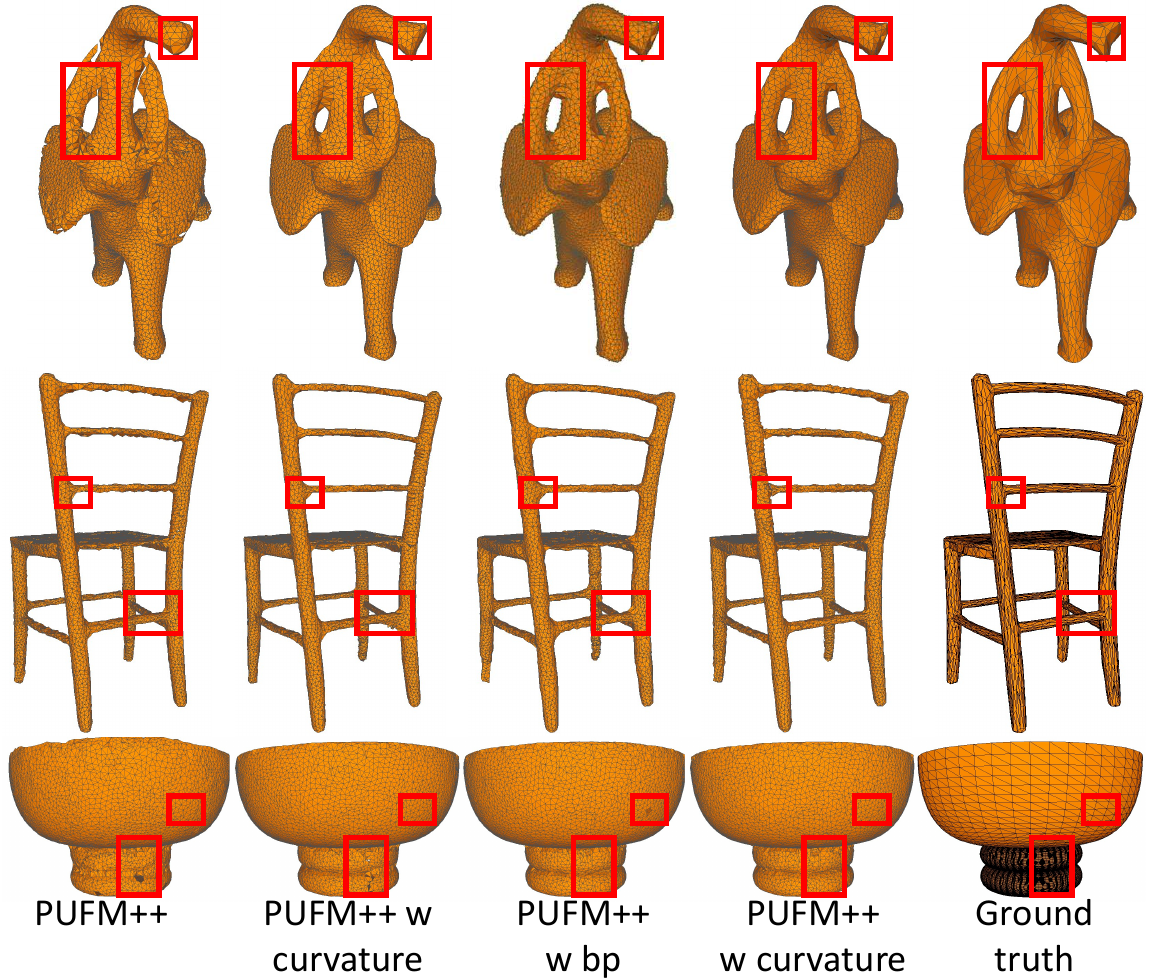}}
		\caption{\textbf{Visual comparison of manifold constraints for mesh reconstruction.} We apply manifold constraints to the PUFM++ inference stage (4-times upsampling) and compare the upsampling quality based on the mesh reconstruction.
		}
		\label{fig:manifold_inference}
\end{figure}

\textbf{Manifold constraints for mesh reconstruction.} In Section~\ref{sec:method}, we introduce the curvature estimation and back projection-based post-processing to adjust the upsampled point clouds. The goal is to use these two technologies to explore a point distribution that better approximates the underlying 3D manifold. They can be useful for downstream applications like mesh reconstruction. Hence, we compare the proposed PUFM++ to its curvature-estimation and back-projection variants and present the results in Figure~\ref{fig:manifold_inference}.

\begin{table}[t]
\caption{\textbf{Ablation on network selection.} CD, HD, and P2F scores on PUGAN and PU1K datasets.}

\centering
\renewcommand\arraystretch{1.7}
\resizebox{\linewidth}{!}{
\begin{tabular}{cc|cccc|cccc} \toprule
\multicolumn{2}{c|}{Inference constraints} & \multicolumn{4}{c|}{PUGAN} & \multicolumn{4}{c}{PU1K} \\
Curvature Est. & Back Proj. & CD & HD & ALR & NC & CD & HD & ALR & NC \\ \midrule
 &  & 0.286 & 0.687 & 0.250 & 0.957 & 0.176 & 0.429 & 0.267 & 0.950 \\
$\checkmark$ &  & 0.291 & 0.648 & 0.242 & 0.988 & 0.181 & 0.424 & 0.246 & 0.986 \\
 & $\checkmark$ & 0.284 & 0.651 & 0.280 & 0.989 & 0.176 & 0.423 & 0.289 & 0.988 \\
\cellcolor{mistyrose}{$\checkmark$} & \cellcolor{mistyrose}{$\checkmark$} & \cellcolor{mistyrose}{0.281} & \cellcolor{mistyrose}{0.650} & \cellcolor{mistyrose}{0.282} & \cellcolor{mistyrose}{0.990} & \cellcolor{mistyrose}{0.175} & \cellcolor{mistyrose}{0.422} & \cellcolor{mistyrose}{0.290} & \cellcolor{mistyrose}{0.990} \\ \bottomrule
\end{tabular}%
}
\label{tab:ablation_constraints}
\end{table}

\begin{figure*}[t]
	\centering
		\centerline{\includegraphics[width=\textwidth]{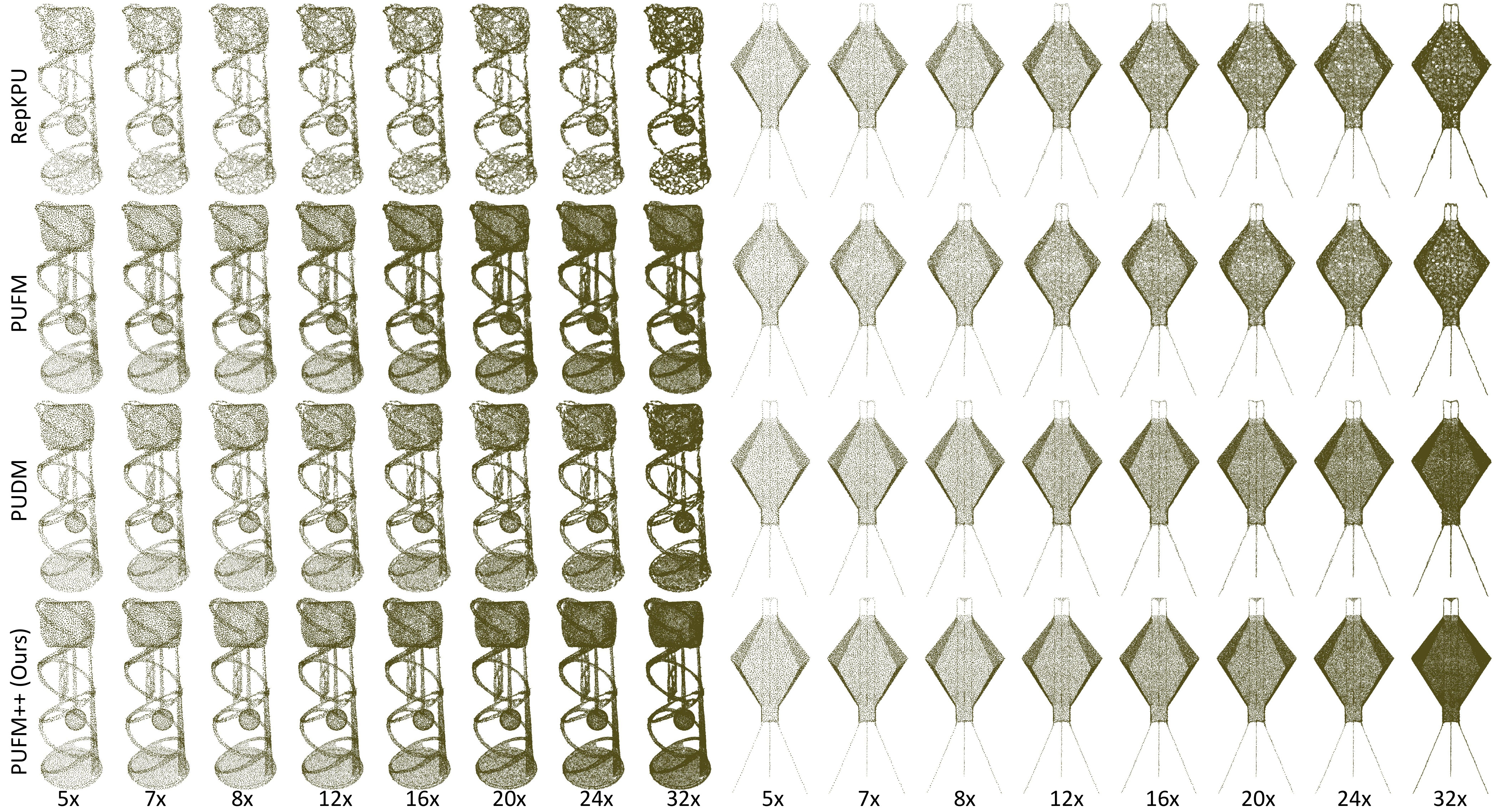}}
		\caption{\textbf{Visual comparison of multiscale point cloud upsampling.} We apply multiscale upsampling on two examples and compare the upsampling quality using different methods.
		}
		\label{fig:multiscale}
\end{figure*}

We reconstructed meshes from the upsampled point clouds using the Ball Pivoting algorithm in MeshLab and visualized the differences. As shown, using PUFM++ alone does not yield satisfactory mesh reconstruction results. In contrast, incorporating back projection or curvature estimation significantly improves mesh quality. For example, in the first row, the nose and teeth of the elephant are much more clearly restored compared to the baseline PUFM++ output. In the highlighted red box regions of the chair and bowl, both curvature estimation and back projection enhance the sharp edges, although some noise is introduced. The combination of curvature estimation and back projection achieves the best balance between smoothness and sharpness in mesh reconstruction.

Meanwhile, the quantitative comparisons on the PUGAN and PU1K datasets are presented in Table~\ref{tab:ablation_constraints}. The results demonstrate that incorporating curvature estimation and back projection not only improves point cloud upsampling performance (as reflected by lower CD and HD scores) but also enhances mesh reconstruction quality, evident by higher ALR and NC scores.

\section{Analysis of PUFM++}
\label{sec:analysis}
In this section, we provide an extended analysis on the robustness and generalization of the proposed method.

\textbf{Multiscale point cloud upsampling.} The proposed PUFM++ can also be used for multiscale point cloud upsampling. As illustrated in other methods~\cite{pu-gcn,pudm,pugan,mpu}, we can iteratively apply the model to the sparse point clouds for $4\times$ upsampling, followed by FPS to downsample the points to the desired size. This iterative procedure enables arbitrary-scale upsampling. The visualization of multiscale upsampling is shown in Figure~\ref{fig:multiscale}. We compare our approach with RepKPU, PUDM and PUFM at factors $5,7,8,12,16,20,24$ and $32$. As shown in the first example, PUFM++ consistently reconstructs objects with smoother surfaces and fewer holes. In the second example, our method accurately preserves the thin and straight structures of the legs of the device without introducing noise.

\begin{figure}[t]
	\centering
		\centerline{\includegraphics[width=\columnwidth]{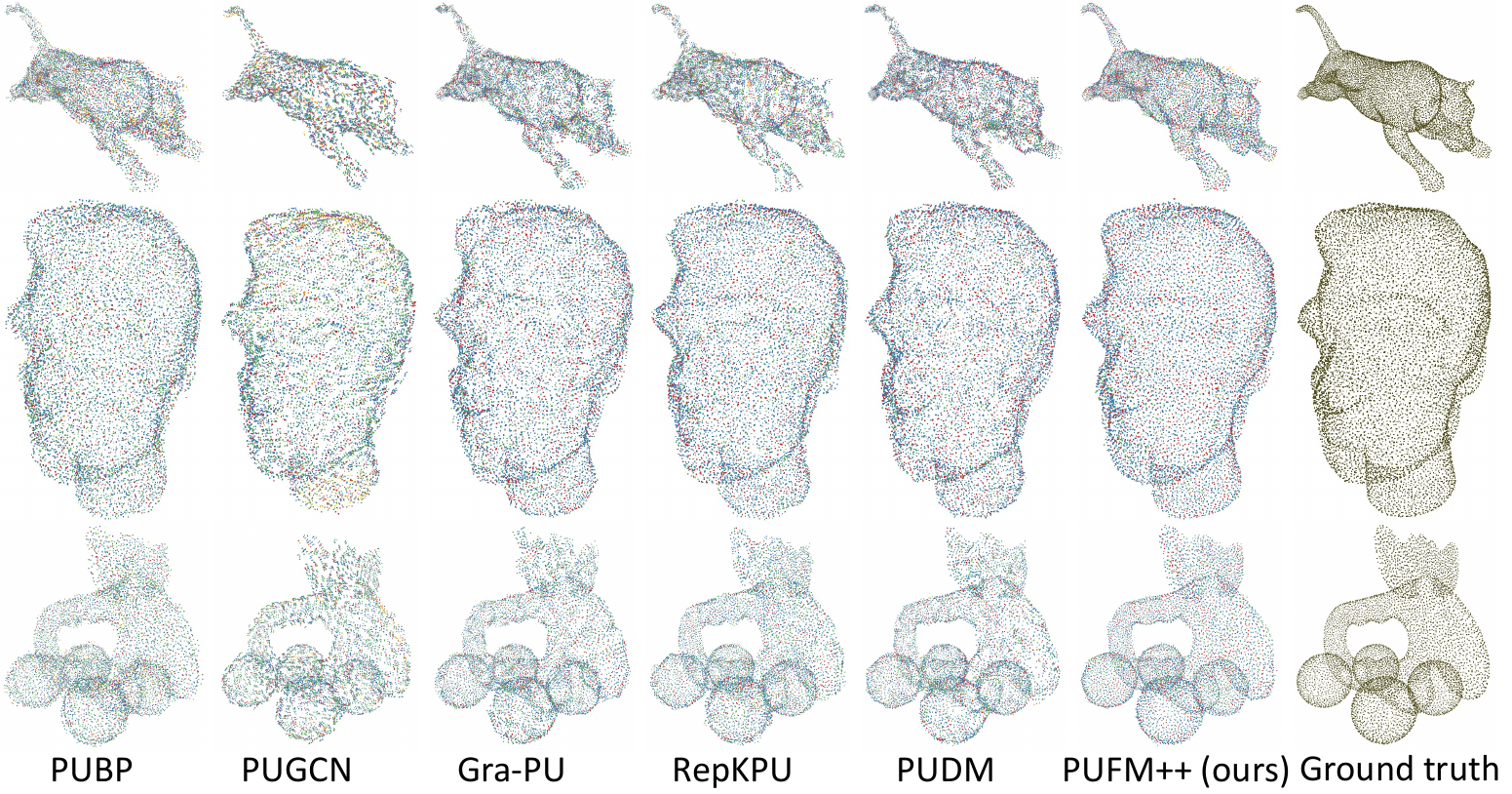}}
		\caption{\textbf{Visual comparison of noisy point cloud upsampling.} We apply point cloud upsampling on PUGAN datasets with noise level $\eta=0.01$ and show three examples for comparison.
		}
		\label{fig:noise}
\end{figure}

\textbf{Noisy point cloud upsampling.} We consider that real-world point clouds suffer from various noise. To test the robustness of our method, we quantitatively compare it with others on noisy point cloud upsampling. We randomly add Gaussian noise to the sparse point clouds with different levels $\eta$, then we use different methods to upsample noisy sparse point clouds to simulate real-world point clouds. We can see from Table~\ref{tab:noise} that our method achieves the lowest CD, HD, and P2F scores among all methods, with the only exception of HD at noise level $\eta=0.01$ for PUDM. Figure~\ref{fig:noise} further illustrates that our upsampled point clouds are visually much smoother than those produced by other methods, effectively mitigating the noise perturbations.

\begin{table}[t]
\centering
\renewcommand\arraystretch{1.3}
\resizebox{\linewidth}{!}{
\begin{tabular}{c|ccc|ccc}
\toprule
Noise level & \multicolumn{3}{c|}{$\eta=$0.01} & \multicolumn{3}{c}{$\eta=$0.02} \\ \cline{2-7} 
Method & CD & HD & P2F & CD & HD & P2F \\ \cline{2-7} 
PUDM & 0.983 & \textbf{1.069} & 5.188 & 2.175 & 2.737 & 1.146 \\
RepKPU & 1.232 & 1.212 & 6.654 & 2.888 & 3.638 & 1.336 \\
Grad-PU & 1.125 & 1.258 & 6.214 & 2.475 & 3.375 & 1.247 \\
PUFM & 0.890 & 1.081 & 5.166 & 1.761 & 2.287 & 1.081 \\
\cellcolor{mistyrose}{Ours} & \cellcolor{mistyrose}{\textbf{0.706}} & \cellcolor{mistyrose}{1.082} & \cellcolor{mistyrose}{\textbf{5.142}} & \cellcolor{mistyrose}{\textbf{1.662}} & \cellcolor{mistyrose}{\textbf{2.231}} & \cellcolor{mistyrose}{\textbf{1.041}} \\ \bottomrule
\end{tabular}%
}
\caption{\textbf{4-times point cloud upsampling on noisy PU1K.} We test on 4-times upsampling and we can see that ours outperforms others, especially on noise level $\eta=0.02$.}
\label{tab:noise}
\end{table}


\begin{table}[t]
\centering
\renewcommand\arraystretch{1.5}
\resizebox{\linewidth}{!}{
\begin{tabular}{c|cccccccc}
\toprule
\begin{tabular}[c]{@{}c@{}}bbox \\ AP@0.7,0.7,0.7\end{tabular} & Sparse PC & Gra-PU & PUDM & PDANS$^{\dagger}$ & PUGAN & PUFM & \cellcolor{mistyrose}{PUFM++} & \begin{tabular}[c]{@{}c@{}}Ground\\ truth\end{tabular} \\ \midrule
Car & 20.35 & 20.50 & 20.18 & 20.25 & 16.78 & 4.22 & \cellcolor{mistyrose}{23.86} & 20.36 \\
Pedestrain & 17.05 & 16.67 & 15.79 & 16.04 & 14.68 & 13.51 & \cellcolor{mistyrose}{16.88} & 18.18 \\ \bottomrule
\end{tabular}%
}
\caption{\textbf{Quantitative metric of the LiDAR data detection.} We test on 2-times upsampling using different methods, and use the upsampled LiDAR data to estimate the object detection performance.}
\label{tab:lidar}
\end{table}

\begin{table}[t]
\centering
\renewcommand\arraystretch{1.5}
\resizebox{\linewidth}{!}{
\begin{tabular}{c|ccccccc}
\toprule
Metric & PUBP & PUGAN & Gra-PU & PUDM & PDANS$^{\dagger}$ & PUFM & \cellcolor{mistyrose}{PUFM++} \\ \midrule
CD & 1.514 & 1.516 & 1.039 & 1.038 & 1.041 & 1.020 & \cellcolor{mistyrose}{0.899} \\
HD & 0.385 & 0.367 & 0.248 & 0.258 & 0.256 & 0.223 & \cellcolor{mistyrose}{0.156}\\ \bottomrule
\end{tabular}%
}
\caption{\textbf{Quantitative metric of the ScanNet point cloud upsampling.} We randomly downsa4-times$4\times$ upsampling task.}
\label{tab:scannet}
\end{table}

\begin{figure*}[t]
	\centering
		\centerline{\includegraphics[width=\textwidth]{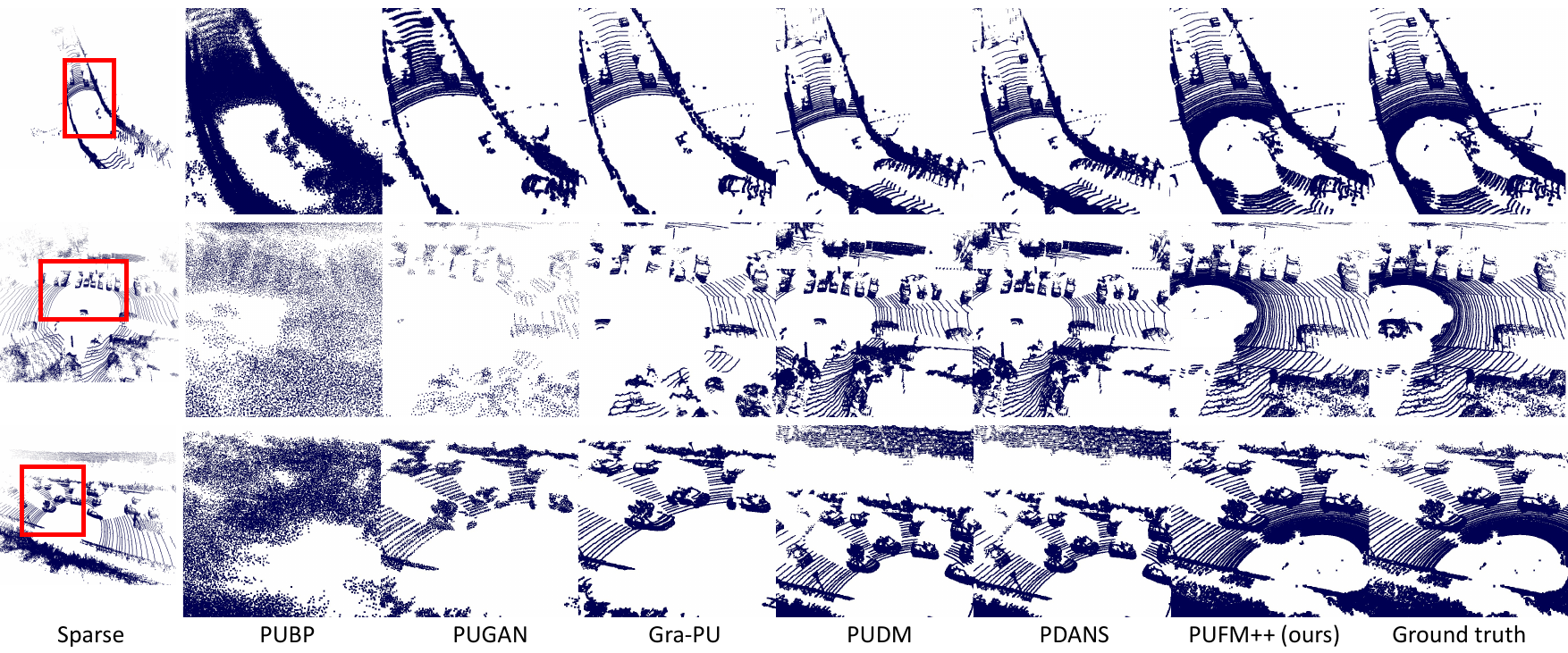}}
		\caption{\textbf{Visual comparison of KITTI LiDAR point cloud upsampling.} We apply different upsampling methods to sparse point clouds for 2$\times$ upsampling and zoom in the red box for visual comparison.
		}
		\label{fig:kitti}
\end{figure*}

\begin{figure*}[t]
	\centering
		\centerline{\includegraphics[width=\textwidth]{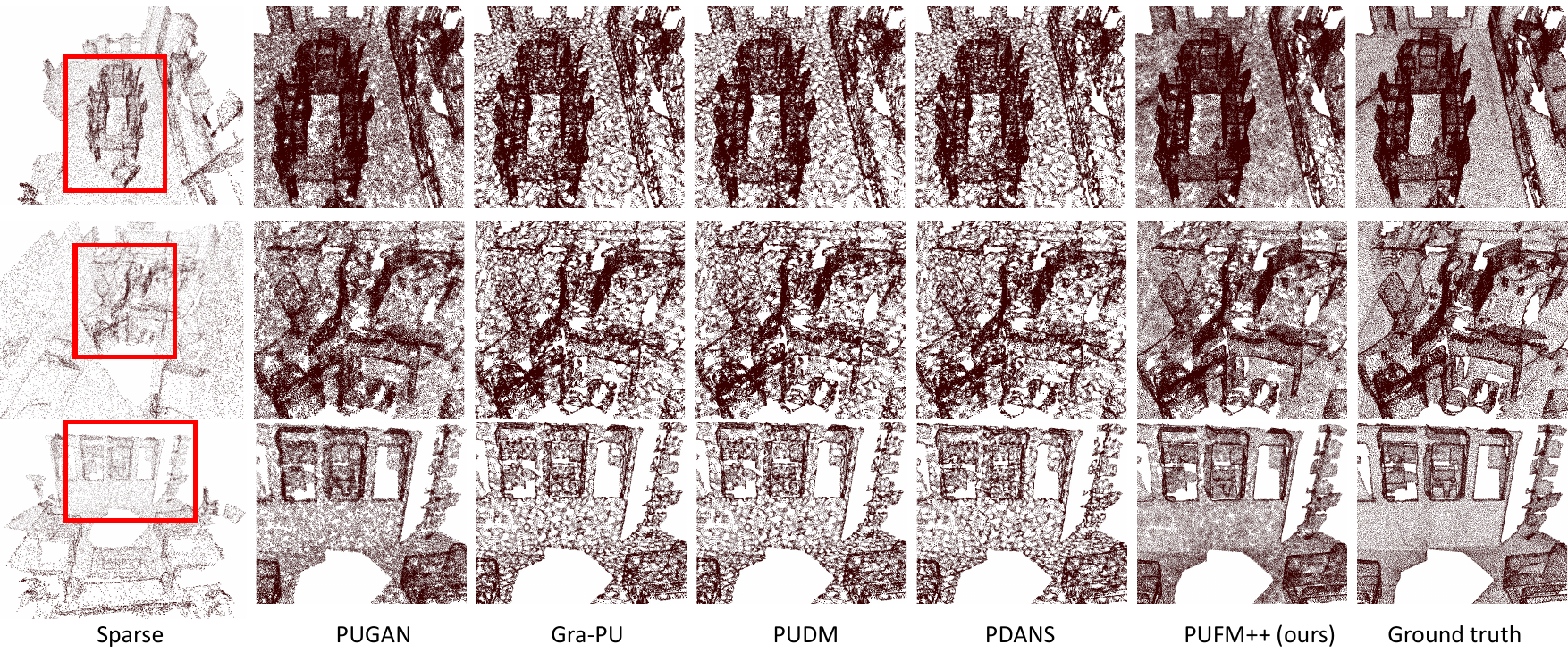}}
		\caption{\textbf{Visual comparison of ScanNet RGB-D point cloud upsampling.} We apply different upsampling methods to sparse point clouds for 4$\times$ upsampling and zoom in the red box for visual comparison.
		}
		\label{fig:scannet}
\end{figure*}

\begin{figure*}[t]
	\centering
		\centerline{\includegraphics[width=\textwidth]{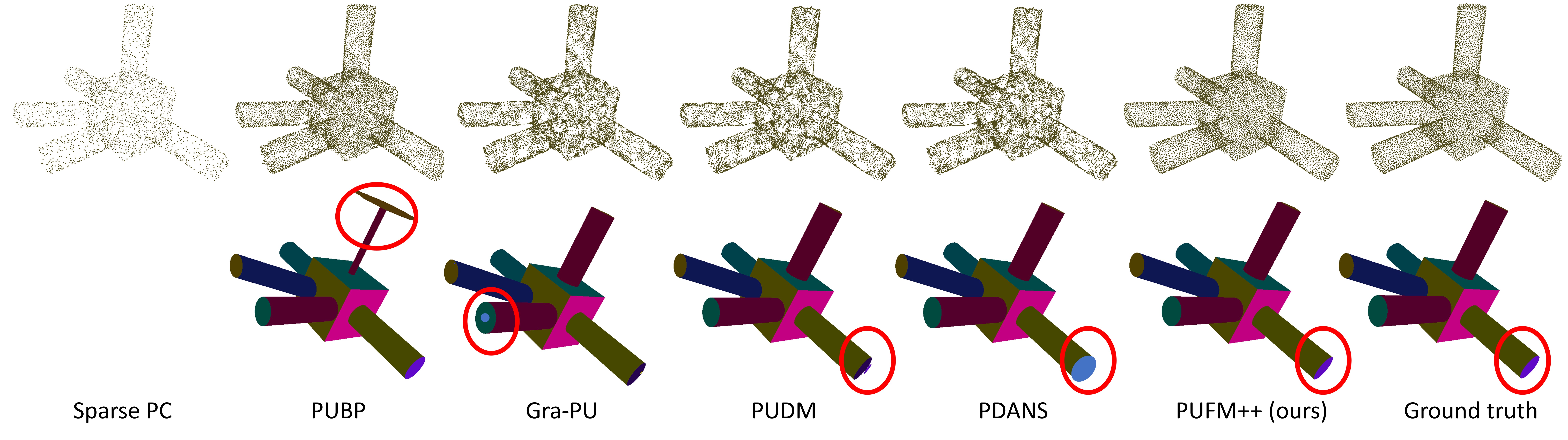}}
		\caption{\textbf{Visual comparison of converting point clouds to CAD models.} We apply different upsampling methods to sparse point clouds for 4$\times$ upsampling and then use Point2CAD to convert them to CAD models. Different colors denote different topological surfaces.
		}
		\label{fig:cad}
\end{figure*}

\begin{figure*}[t]
	\centering
		\centerline{\includegraphics[width=\textwidth]{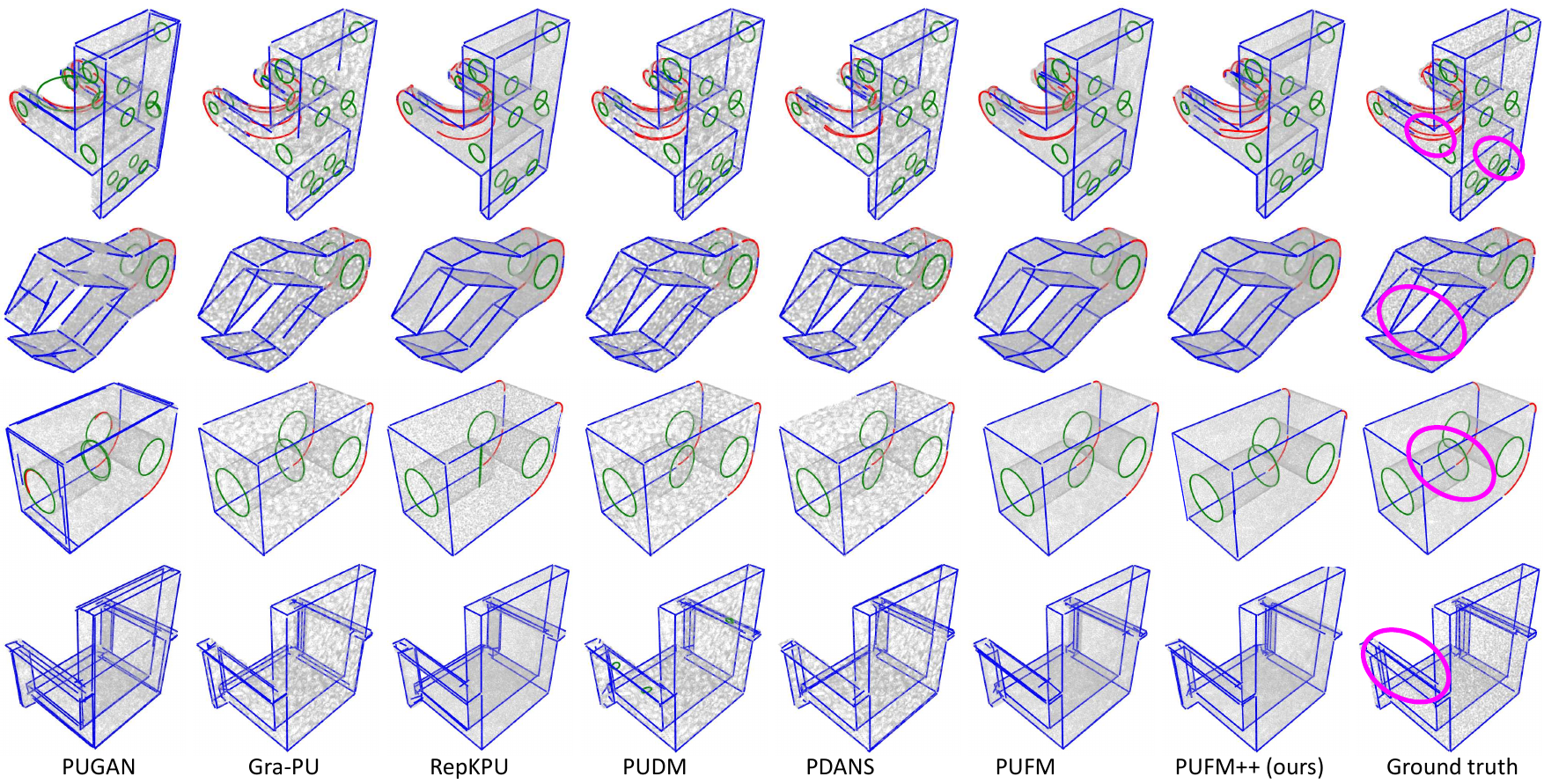}}
		\caption{\textbf{Visual comparison of point clouds curve estimation.} We apply different upsampling methods to sparse point clouds for 4-times upsampling and then use PI3DETR to estimate the 3D curves from point clouds. Different colors denote different edges: red for arcs, blue for lines, green for circles.
		}
		\label{fig:edge}
\end{figure*}

\begin{table}[t]
\centering
\renewcommand\arraystretch{1.5}
\resizebox{\linewidth}{!}{
\begin{tabular}{c|cccccccc}
\toprule \toprule
\multicolumn{9}{c}{Point cloud to CAD} \\ \midrule \midrule
Metric & PUBP & Gra-PU & PUDM & PDANS & RepKPU & PUFM & \cellcolor{mistyrose}{PUFM++} & \begin{tabular}[c]{@{}c@{}}Ground\\ truth\end{tabular} \\ \midrule
Res-err$\downarrow$ & 0.022 & 0.006 & 0.008 & 0.008 & 0.010 & 0.006 & \cellcolor{mistyrose}{0.005} & 0.004 \\
P-cover$\uparrow$ & 0.934 & 0.993 & 0.989 & 0.990 & 0.988 & 0.994 & \cellcolor{mistyrose}{0.995} & 0.997 \\ \midrule \midrule
\multicolumn{9}{c}{3D curve estimation} \\ \midrule \midrule
Metric & PUGAN & Gra-PU & RepKPU & PUDM & PDANS & PUFM & \cellcolor{mistyrose}{PUFM++} & \begin{tabular}[c]{@{}c@{}}Ground\\ truth\end{tabular} \\ \midrule
CD$\downarrow$ & 0.0189 & 0.0040 & 0.0041 & 0.0049 & 0.0045 & 0.0042 & \cellcolor{mistyrose}{0.0038} & 0.0032 \\
HD$\downarrow$ & 0.156 & 0.0674 & 0.0675 & 0.0678 & 0.0676 & 0.0674 & \cellcolor{mistyrose}{0.0670} & 0.0665 \\ \bottomrule \bottomrule
\end{tabular}%
}
\caption{\textbf{Quantitative metric of the point cloud for CAD analysis.} We test on 4$\times$ upsampling on two point clouds to CAD applications: 1) point cloud to CAD conversion and 2) 3D curve estimation.}
\label{tab:cad}
\end{table}

\textbf{Real-world point cloud upsampling.}
To demonstrate the generalization of the proposed method, we apply it to real-world point clouds, including ABC~\cite{abc} (computer-aided design (CAD) dataset), ScanNet~\cite{scannet} (RGB-D video dataset) and KITTI~\cite{kitti} (LiDAR driving dataset). We consider both upsampling quality and performance on downstream tasks, including object detection and segmentation.

For real-world point cloud upsampling, we directly apply pretrained models to ScanNet and KITTI. For ScanNet, we select 20 samples and randomly downsample each by $4$-times to obtain sparse point clouds. We then apply different upsampling methods to recover dense point clouds for comparison. 

For KITTI, we simulate 32-beam LiDAR scans from the original 64-beam point clouds using angular sampling. Each sparse point cloud is then upsampled by $4$-times, followed by a $2$-times downsampling using FPS. The resulting point clouds are fed into SECOND \cite{yan2018second} for 3D object detection. We report the average precision of the bird's-eye view bounding boxes at the easy difficulty level in Table~\ref{tab:lidar}. Our method achieves the highest AP among all competing methods. Remarkably, it even outperforms the original LiDAR data on the car class. Figure~\ref{fig:kitti} provides qualitative comparisons on KITTI, with red boxes highlighting regions of interest. Our method preserves beam-like LiDAR patterns similar to the original data. In contrast, PUBP produces noticeable blurring across the scene, while PUGAN and Gra-PU fail to generate consistent beam structures. PUDM and PDANS yield comparable results but introduce misalignment in fine details.

For the ScanNet dataset, we randomly sample 10 scenes and downsample each point cloud by a factor of $4$-times to generate sparse inputs. We then apply various upsampling methods to recover the dense point clouds with the same $4$-times upsampling ratio. To quantitatively assess the reconstruction quality, we report the CD and HD scores in Table~\ref{tab:scannet}. We can see that our method achieves the best CD and HD scores. We also provide qualitative comparisons in Figure~\ref{fig:scannet}. As highlighted in the zoom-in regions (red boxes), our method more accurately restores fine-grained 3D surfaces, producing smooth and uniformly distributed points, like the floor in all three examples, and the cupboard in the third example. In contrast, competing approaches often struggle to reconstruct the underlying geometry and tend to generate point clouds with noticeable holes or irregularities.

To validate the robustness of our point cloud upsampling approach on industrial manufacturing, we apply our model to CAD object recognition, including 1) point to CAD conversion, and 2) 3D curve estimation. 

For task 1), we randomly select 10 point clouds from the ABC datasets with their corresponding label information. We randomly downsample the original point clouds by 4-times and then upsample them by using different upsampling methods. Finally, we use kNN tree search to align the upsampled points with the original labels. Next, we use Point2CAD~\cite{point2cad}, a recent work on reconstructing CAD models from raw point clouds, to convert upsampled point clouds to CAD models. We visualize one example in Figure~\ref{fig:cad}. The first row includes the upsampled point clouds, and the second row the reconstructed CAD models, with different colors denoting different topological surfaces. From the marked red circles, we can see that our proposed method produces the closest CAD model to the ground truth. 

For task 2), we use the same set of 10 point clouds from the ABC dataset to produce sparse and original dense point clouds. Our goal is to estimate the 3D curves of the point clouds to see if the upsampling method can preserve the 3D curvatures. We apply the recent parametric instance detection of 3D point cloud edges, PI3DETR~\cite{pi3detr}, to the point clouds and visualize the estimated curves. In Figure~\ref{fig:edge}, we show the edge estimation with different colors. To highlight the differences, we circle them in pink. We can see that 1) our method correctly produces the green circles in the first and third example, as well as the straight lines in the second and fourth examples; 2) Visually, PUGAN and Grad-PU produce more extra or incomplete edges; 3) PUDM produces extra circles in the fourth example; 4) PDANS and PUFM are closer to our approach, but they also fail at fine details.

For both tasks (1) and (2), we quantitatively compare different upsampling methods using several evaluation metrics. For task (1), we adopt Res-Err to measure the discrepancy between the reconstructed surface and the ground-truth surface, where the correspondence is established using Hungarian matching. We additionally report P-Cover, which quantifies the proportion of the input point cloud that is covered by the generated surface (see Point2CAD~\cite{point2cad} for metric definitions). Our method achieves the second-best performance, only slightly lower than that obtained using the ground-truth point clouds. For task (2), we evaluate reconstruction quality in edge regions using Chamfer Distance (CD) and Hausdorff Distance (HD), where lower values indicate better geometric consistency. Except for the ground truth data, our approach achieves the best performance among all competing upsampling methods.

\section{Ethical Discussion}

\paragraph{Practical Impact.}
PUFM++ provides a unified and highly efficient framework for point cloud upsampling with applications in 3D vision, robotics, AR/VR, reverse engineering, and digital twins.
First, PUFM++ can help practitioners reconstruct high-fidelity 3D geometry from sparse, noisy, or incomplete scans, enabling more accessible digitization of cultural heritage, industrial assets, and environmental data.
Second, the method can support robotics and autonomous systems by improving scene understanding, particularly for manipulation and navigation in cluttered or partially observed environments.
Finally, PUFM++ may facilitate content creation for film, gaming, and 3D design tools, reducing manual labor for artists and engineers.

However, the ability to enhance sparse 3D data can introduce risks. PUFM++ could be misused to artificially densify or restore 3D scans in ways that obscure provenance or authenticity. This may enable sophisticated forms of 3D spoofing, for example, producing deceptive digital replicas of individuals, objects, or scenes. As with any generative AI system, responsible use requires transparency about whether 3D content has been algorithmically modified or reconstructed.

\paragraph{Societal Impact.}
PUFM++ is trained on benchmark datasets that primarily contain synthetic or curated 3D models, which do not comprehensively represent global object distributions, environmental diversity, or scanning conditions. As a result, the model may generalize unevenly to real-world scans that contain domain shifts such as culturally specific artifacts, region-specific object designs, or sensor-specific distortions.
Moreover, because upsampling methods implicitly learn priors over shape structures, they may unintentionally impose biases, e.g., over-smoothing culturally distinctive geometry or failing on underrepresented categories. Such biases can propagate into downstream tasks like mesh reconstruction, robotics perception, or 3D mapping, potentially affecting reliability in safety-critical systems.

PUFM++ does not explicitly model personal identity or biometric information, but high-fidelity reconstruction of human-related geometry (e.g., faces, bodies) could raise privacy considerations if used on sensitive scans without consent. Users should carefully evaluate the appropriateness of applying generative reconstruction models in domains involving personal or proprietary 3D data.

\paragraph{Environmental Impact.}
Our experiments rely on 2 NVIDIA V100 GPUs, each with an approximate power draw of 300W (V100) during training. Across all experiments, we required roughly 800 GPU hours. Training the full PUFM++ model requires around 10 GPU hours on a V100 GPU, corresponding to approximately 3 kWh of electricity and an estimated 1.2kg of CO$_2$ emissions, depending on local energy sources. While the training cost is modest compared to large-scale generative models, we encourage the community to adopt energy-efficient practices such as mixed-precision training, model reuse through fine-tuning, and sharing pretrained models to reduce redundant computations.

\section{Conclusions}
\label{sec:conclusions}
We introduce PUFM++, a versatile point cloud upsampling framework built upon a novel two-stage flow-matching strategy, combined with inference-time optimization, to achieve state-of-the-art performance for arbitrary upsampling factors. Our results show that the proposed two-stage flow-matching process significantly enhances geometric restoration and provides strong robustness against noise. The inference-time optimization further improves alignment with the underlying 3D manifold. To the best of our knowledge, we are also the first to adopt a recurrent inference network for point cloud processing, which consistently outperforms PointNet++ across all evaluation metrics.
Extensive analyses and visualizations validate our findings and demonstrate the superior performance of PUFM++. We also compare our model with existing approaches across multiple downstream tasks and various real-world datasets, including KITTI, ScanNet, and ABC, highlighting the robustness and efficiency of our method.
Future work includes one-step flow-matching distillation and joint upsampling–completion models, aiming to further improve efficiency and generalization.

\vspace{5mm}
\noindent \textbf{Funding information.} Z.~S.~Liu has been supported by the Research Council of Finland's decision number 359633.
Chenhang He acknowledges support from the National Natural Science Foundation of China (No.62406268).
R.~Maier acknowledges support from the German Academic Exchange Service (DAAD), project ID 57711336.

\noindent \textbf{Conflict of interest statement.} The authors declare that they have no conflict of interest.

\noindent \textbf{Data availability.} Code and pretrained models are publicly available at {\small{\url{https://github.com/Holmes-Alan/Enhanced_PUFM}}}

\bibliographystyle{splncs04}
\bibliography{sec/shortstrings,references}

@String(CVPR= {IEEE Conf. Comput. Vis. Pattern Recog.})

@String(ICCV= {Int. Conf. Comput. Vis.})

@String(ECCV= {Eur. Conf. Comput. Vis.})

@String(NIPS= {Adv. Neural Inform. Process. Syst.})

@String(TVCG  = {IEEE Trans. Vis. Comput. Graph.})

@String(ICIP = {IEEE Int. Conf. Image Process.})

@String(ACCV  = {ACCV})

@String(ICLR = {Int. Conf. Learn. Represent.})

@String(AAAI = {AAAI})

@article{yan2018second,
  title={SECOND: Sparsely Embedded Convolutional Detection},
  author={Yan, Yan and Mao, Yuxing and Li, Bo},
  journal={Sensors},
  volume={18},
  number={10},
  pages={3337},
  year={2018},
  publisher={MDPI},
}

@inproceedings{robot,
  title={Spot-Compose: A Framework for Open-Vocabulary Object Retrieval and Drawer Manipulation in Point Clouds},
  author={Oliver Lemke and et al.},
  booktitle={Int. Conf. on Robotics and Automation},
  year={2024},
}

@inproceedings{car,
  title={Visual Point Cloud Forecasting enables Scalable Autonomous Driving},
  author={Yang, Zetong and et al.},
  booktitle=CVPR,
  year={2024}
}

@article{p2m,
  title = {Point2Mesh: A Self-Prior for Deformable Meshes},
  author = {Hanocka, Rana and et al.},
  year = {2020},
  issue_date = {July 2020}, 
  volume = {39}, 
  number = {4}, 
  issn = {0730-0301},
  journal = {ACM Trans. Graph.}, 
}

@article{punet,
	author = {L. Yu and et al.},
	journal = CVPR,
	title = {PU-Net: Point Cloud Upsampling Network},
	year = {2018},
	volume = {},
	issn = {},
	pages = {2790-2799},
	address = {Los Alamitos, CA, USA},
}

@article{ecnet,
	author = {Yu, Lequan and et al.},
	title = {EC-Net: an Edge-aware Point set Consolidation Network},
	journal = ECCV,
	year = {2018}
}

@article{mpu,
	author = {W. Yifan and et al.},
	journal = CVPR,
	title = {Patch-Based Progressive 3D Point Set Upsampling},
	year = {2019},
	pages = {5951-5960},
	address = {Los Alamitos, CA, USA},
}

@article{pugan,
	author = {R. Li and X. Li and C. Fu and D. Cohen-Or and P. Heng},
	journal = ICCV,
	title = {PU-GAN: A Point Cloud Upsampling Adversarial Network},
	year = {2019},
	pages = {7202-7211},
	address = {Los Alamitos, CA, USA},
}

@inproceedings{pugeo,
author = {Qian, Yue and Hou, Junhui and Kwong, Sam and He, Ying},
title = {PUGeo-Net: A Geometry-Centric Network for 3D Point Cloud Upsampling},
year = {2020},
booktitle = {Computer Vision – ECCV 2020: 16th European Conference, Glasgow, UK, August 23–28, 2020, Proceedings, Part XIX},
pages = {752–769},
numpages = {18},
location = {Glasgow, United Kingdom}
}

@INPROCEEDINGS{pointnet,
	author = {R. Charles and H. Su and M. Kaichun and L. J. Guibas},
	booktitle = CVPR,
	title = {PointNet: Deep Learning on Point Sets for 3D Classification and Segmentation},
	year = {2017},
	volume = {},
	pages = {77-85},
	address = {Los Alamitos, CA, USA},
}

@INPROCEEDINGS{pointconv,
  author={Wu, Wenxuan and Qi, Zhongang and Fuxin, Li},
  title={Deep Convolutional Networks on 3D Point Clouds},
  booktitle= CVPR, 
  year={2019},
  volume={},
  number={},
  pages={9613-9622},
}

@inproceedings{point_v2,
author = {Wu, Xiaoyang and Lao, Yixing and Jiang, Li and Liu, Xihui and Zhao, Hengshuang},
title = {Point transformer V2: grouped vector attention and partition-based pooling},
year = {2022},
booktitle = NIPS,
articleno = {2415},
numpages = {13},
}

@inproceedings{point_v3,
    title={Point Transformer V3: Simpler, Faster, Stronger},
    author={Wu, Xiaoyang and Jiang, Li and Wang, Peng-Shuai and Liu, Zhijian and Liu, Xihui and Qiao, Yu and Ouyang, Wanli and He, Tong and Zhao, Hengshuang},
    booktitle=CVPR,
    year={2024}
}

@article{dgcnn,
	author = {Wang, Yue and et al.},
	title = {Dynamic Graph CNN for Learning on Point Clouds},
	year = {2019},
	issue_date = {November 2019},
	address = {New York, NY, USA},
	volume = {38},
	number = {5},
	journal = {ACM Trans. Graph.},
	articleno = {146},
	numpages = {12},
}

@article{adam,
	author = {Kingma, Diederik P. and Ba, Jimmy},
	journal = ICLR,
	title = {Adam: A Method for Stochastic Optimization},
	year = 2014
}

@ARTICLE{kitti,
	author = {Andreas Geiger and et al.},
	title = {Vision meets Robotics: The KITTI Dataset},
	journal = {International Journal of Robotics Research (IJRR)},
	year = {2013}
}

@INPROCEEDINGS{abc,
  author={Koch, Sebastian and Matveev, Albert and Jiang, Zhongshi and Williams, Francis and Artemov, Alexey and Burnaev, Evgeny and Alexa, Marc and Zorin, Denis and Panozzo, Daniele},
  booktitle=CVPR, 
  title={ABC: A Big CAD Model Dataset for Geometric Deep Learning}, 
  year={2019},
  volume={},
  number={},
  pages={9593-9603},
}

@inproceedings{dispu,
     title={Point Cloud Upsampling via Disentangled Refinement},
     author={Li, Ruihui and et al.},
     booktitle = CVPR,
     year = {2021}
 }

@ARTICLE{pufm,
  author={Zhi-Song Liu and Chenhang He and Yakun Ju and Lei Li},
  journal=AAAI, 
  title={{PUFM}: Efficient Point Cloud Upsampling via Flow Matching}, 
  year={2026},
}

@ARTICLE{puflow,
  author={Mao, Aihua and Du, Zihui and Hou, Junhui and Duan, Yaqi and Liu, Yong-Jin and He, Ying},
  journal=TVCG, 
  title={PU-Flow: A Point Cloud Upsampling Network With Normalizing Flows}, 
  year={2023},
  volume={29},
  number={12},
  pages={4964-4977},
}

@InProceedings{pu-gcn,
    author    = {Qian, Guocheng and et al.},
    title     = {PU-GCN: Point Cloud Upsampling Using Graph Convolutional Networks},
    booktitle = CVPR,
    year      = {2021},
    pages     = {11683-11692}
}

@InProceedings{dualbp,
  author={Liu, Zhi-Song and et al.},
  booktitle=ICIP, 
  title={Arbitrary Point Cloud Upsampling Via Dual Back-Projection Network}, 
  year={2023},
  volume={},
  number={},
  pages={1470-1474}
}

@InProceedings{pcflow,
    author    = {Wu, Lemeng and Wang, Dilin and Gong, Chengyue and Liu, Xingchao and Xiong, Yunyang and Ranjan, Rakesh and Krishnamoorthi, Raghuraman and Chandra, Vikas and Liu, Qiang},
    title     = {Fast Point Cloud Generation With Straight Flows},
    booktitle = CVPR,
    year      = {2023},
    pages     = {9445-9454}
}

@InProceedings{grad-pu,
  author={He, Yun and et al.},
  booktitle=CVPR, 
  title={Grad-PU: Arbitrary-Scale Point Cloud Upsampling via Gradient Descent with Learned Distance Functions}, 
  year={2023},
  volume={},
  number={},
  pages={5354-5363}
}

@INPROCEEDINGS{pdm,
  author={Luo, Shitong and Hu, Wei},
  booktitle=CVPR, 
  title={Diffusion Probabilistic Models for 3D Point Cloud Generation}, 
  year={2021},
  volume={},
  number={},
  pages={2836-2844},
}

@article{GFNet,
title={{GFN}et: Geometric Flow Network for 3D Point Cloud Semantic Segmentation},
author={Haibo Qiu and Baosheng Yu and Dacheng Tao},
journal={Trans. Mac. Lear. Res.},
issn={2835-8856},
year={2022},
}

@InProceedings{pudm,
  author={Qu, Wentao and et al.},
  booktitle=CVPR, 
  title={A Conditional Denoising Diffusion Probabilistic Model for Point Cloud Upsampling}, 
  year={2024},
  volume={},
  number={},
  pages={20786-20795}
}

@inproceedings{point2cad,
  title={Point2CAD: Reverse Engineering CAD Models from 3D Point Clouds},
  author={Liu, Yujia and Obukhov, Anton and Wegner, Jan Dirk and Schindler, Konrad},
  booktitle=CVPR,
  pages={3763--3772},
  year={2024}
}

@article{pi3detr,
      title={PI3DETR: Parametric Instance Detection of 3D Point Cloud Edges with a Geometry-Aware 3DETR}, 
      author={Fabio F. Oberweger and Michael Schwingshackl and Vanessa Staderini},
      year={2025},
      journal={ArXiv Preprint, 2509.03262},
}

@InProceedings{pdans,
    author    = {Zhang, Boqian and Yang, Shen and Chen, Hao and Yang, Chao and Jia, Jing and Jiang, Guang},
    title     = {Point Cloud Upsampling Using Conditional Diffusion Module with Adaptive Noise Suppression},
    booktitle = CVPR,
    year      = {2025},
    pages     = {16987-16996}
}

@article{normflow,
author = {Papamakarios, George and Nalisnick, Eric and Rezende, Danilo Jimenez and Mohamed, Shakir and Lakshminarayanan, Balaji},
title = {Normalizing flows for probabilistic modeling and inference},
year = {2021},
volume = {22},
number = {1},
journal = {J. Mach. Learn. Res.},
articleno = {57},
numpages = {64},
}

@inproceedings{ddpm,
    author = {Ho, Jonathan and et al.},
    title = {Denoising diffusion probabilistic models},
    year = {2020},
    isbn = {9781713829546},
    booktitle = NIPS,
    articleno = {574},
    numpages = {12},
    location = {Vancouver, BC, Canada}
}

@inproceedings{flow,
title={Flow Matching for Generative Modeling},
author={Yaron Lipman and et al.},
booktitle={The Eleventh International Conference on Learning Representations },
year={2023},
}

@article{cfm,
title={Improving and generalizing flow-based generative models with minibatch optimal transport},
author={Alexander Tong and Kilian FATRAS and Nikolay Malkin and Guillaume Huguet and Yanlei Zhang and Jarrid Rector-Brooks and Guy Wolf and Yoshua Bengio},
journal={Transactions on Machine Learning Research},
issn={2835-8856},
year={2024}
}

@inproceedings{sb_1,
title={Diffusion Schr\"odinger Bridge Matching},
author={Yuyang Shi and Valentin De Bortoli and Andrew Campbell and Arnaud Doucet},
booktitle={Thirty-seventh Conference on Neural Information Processing Systems},
year={2023},
}

@article{sb_2,
      title={Stochastic Interpolants: A Unifying Framework for Flows and Diffusions}, 
      author={Michael S. Albergo and Nicholas M. Boffi and Eric Vanden-Eijnden},
      year={2023},
      journal={ArXiv Preprint, 2303.08797},
}

@inproceedings{emd,
    title={PointMixup: Augmentation for Point Clouds}, 
    author={Yunlu Chen and et al.},
    year={2020},
    booktitle=ECCV,
}

@article{gan,
author = {Goodfellow, Ian and Pouget-Abadie, Jean and Mirza, Mehdi and Xu, Bing and Warde-Farley, David and Ozair, Sherjil and Courville, Aaron and Bengio, Yoshua},
title = {Generative adversarial networks},
year = {2020},
volume = {63},
number = {11},
journal = {Commun. ACM},
pages = {139–144},
numpages = {6}
}

@inproceedings{pointnet++,
author = {Qi, Charles R. and et al.},
title = {PointNet++: deep hierarchical feature learning on point sets in a metric space},
year = {2017},
booktitle = NIPS,
pages = {5105–5114},
numpages = {10},
series = {NIPS'17}
}

@inproceedings{repkpu,
  title={RepKPU: Point Cloud Upsampling with Kernel Point Representation and Deformation},
  author={Rong, Yi and et al.},
  booktitle=CVPR,
  pages={21050--21060},
  year={2024}
}

@ARTICLE{pvcn,
  author={Lu, Xuequan and et al.},
  journal=TVCG, 
  title={Low Rank Matrix Approximation for 3D Geometry Filtering}, 
  year={2022},
  volume={28},
  number={4},
  pages={1835-1847}
}

@InProceedings{deepla,
    author    = {Zeng, Ziyin and Dong, Mingyue and Zhou, Jian and Qiu, Huan and Dong, Zhen and Luo, Man and Li, Bijun},
    title     = {DeepLA-Net: Very Deep Local Aggregation Networks for Point Cloud Analysis},
    booktitle = CVPR,
    year      = {2025},
    pages     = {1330-1341}
}

@inproceedings{point,
  title={Point transformer},
  author={Zhao, Hengshuang and Jiang, Li and Jia, Jiaya and Torr, Philip HS and Koltun, Vladlen},
  booktitle=ICCV,
  pages={16259--16268},
  year={2021}
}

@article{kpconv,
    Author = {Thomas, Hugues and et al.},
    Title = {KPConv: Flexible and Deformable Convolution for Point Clouds},
    Journal = ICCV,
    Year = {2019}
}

@ARTICLE{sr_1,
  author={Liu, Zhi-Song and Siu, Wan-Chi and Chan, Yui-Lam},
  journal={IEEE Transactions on Circuits and Systems for Video Technology}, 
  title={Photo-Realistic Image Super-Resolution via Variational Autoencoders}, 
  year={2021},
  volume={31},
  number={4},
  pages={1351-1365}
}

@InProceedings{sr_2,
    author = {Liu, Zhi-Song and Wang, Li-Wen and Li, Chu-Tak and Siu, Wan-Chi},
    title = {Image Super-Resolution via Attention based Back Projection Networks},
    booktitle = {IEEE International Conference on Computer Vision Workshop(ICCVW)},
    year = {2019}
}

@inproceedings{np,
    author    = {Wanquan Feng and et al.},
    title     = {Neural Points: Point Cloud Representation with Neural Fields for Arbitrary Upsampling},
    booktitle = CVPR,
    year      = {2022}
}

@article{spunet,
author = {Liu, Xinhai and Liu, Xinchen and Liu, Yu-Shen and Han, Zhizhong},
title = {SPU-Net: Self-Supervised Point Cloud Upsampling by Coarse-to-Fine Reconstruction With Self-Projection Optimization},
year = {2022},
volume = {31},
issn = {1057-7149},
journal = {Trans. Img. Proc.},
pages = {4213–4226},
numpages = {14}
}

@inproceedings{putransformer,
  title={Pu-transformer: Point cloud upsampling transformer},
  author={Qiu, Shi and Anwar, Saeed and Barnes, Nick},
  booktitle=ACCV,
  pages={2475--2493},
  year={2022}
}

@ARTICLE{pu-dense,
  author={Akhtar, Anique and Li, Zhu and Auwera, Geert Van der and Li, Li and Chen, Jianle},
  journal={IEEE Transactions on Image Processing}, 
  title={PU-Dense: Sparse Tensor-Based Point Cloud Geometry Upsampling}, 
  year={2022},
  volume={31},
  number={},
  pages={4133-4148},
}

@inproceedings{pvd,
    author    = {Zhou, Linqi and Du, Yilun and Wu, Jiajun},
    title     = {3D Shape Generation and Completion Through Point-Voxel Diffusion},
    booktitle = ICCV,
    year      = {2021},
    pages     = {5826-5835}
}

@inproceedings{scannet,
    title={ScanNet: Richly-annotated 3D Reconstructions of Indoor Scenes},
    author={Dai, Angela and Chang, Angel X. and Savva, Manolis and Halber, Maciej and Funkhouser, Thomas and Nie{\ss}ner, Matthias},
    booktitle = {Proc. Computer Vision and Pattern Recognition (CVPR), IEEE},
    year = {2017}
}

@inproceedings{tetsphere,
    title={TetSphere Splatting: Representing High-Quality Geometry with Lagrangian Volumetric Meshes},
    author={Minghao Guo and Bohan Wang and Kaiming He and Wojciech Matusik},
    booktitle={The Thirteenth International Conference on Learning Representations},
    year={2025},
}

@inproceedings{superpc,
  title = {{SuperPC}: A Single Diffusion Model for Point Cloud Completion, Upsampling, Denoising, and Colorization},
  author = {Du, Yi and Zhao, Zhipeng and Su, Shaoshu and Golluri, Sharath and Zheng, Haoze and Yao, Runmao and Wang, Chen},
  booktitle = {IEEE/CVF Conference on Computer Vision and Pattern Recognition (CVPR)},
  year = {2025}
}

@inproceedings{sapcu,
  title = {Self-Supervised Arbitrary-Scale Point Clouds Upsampling via Implicit Neural Representation},
  author = {Wenbo Zhao and Xianming Liu and Zhiwei Zhong and Junjun Jian and Wei Gao and Ge Li and Xiangyang Ji},
  booktitle = {Proceedings IEEE Conf. on Computer Vision and Pattern Recognition (CVPR)},
  year      = {2022},
  pages     = {1999-2007}
}

@inproceedings{rin,
    author = {Jabri, Allan and Fleet, David J. and Chen, Ting},
    title = {Scalable adaptive computation for iterative generation},
    year = {2023},
    booktitle = {Proceedings of the 40th International Conference on Machine Learning},
    articleno = {594},
    numpages = {21},
    location = {Honolulu, Hawaii, USA},
}

@STRING{accv = {Proc. Asian Conf. on Computer Vision}}

@STRING{cvpr = {CVPR}}

@STRING{iccvw = {ICCV-W}}

@STRING{ICCVW = {ICCV-W}}

@STRING{eccv = {ECCV}}

@STRING{iccv = {ICCV}}

@STRING{iclr = {  ICLR  }}

@STRING{icip = {ICIP}}

@STRING{ijrr = {IJRR}}

@STRING{arxiv = {arXiv}}

\end{document}